\documentclass{article}

\usepackage[nonatbib, preprint]{neurips_2022}

\usepackage[utf8]{inputenc} % allow utf-8 input
\usepackage[T1]{fontenc}    % use 8-bit T1 fonts
\usepackage{hyperref}       % hyperlinks
\usepackage{url}            % simple URL typesetting
\usepackage{booktabs}       % professional-quality tables
\usepackage{amsfonts}       % blackboard math symbols
\usepackage{nicefrac}       % compact symbols for 1/2, etc.
\usepackage{microtype}      % microtypography
\usepackage{xcolor}         % colors

\usepackage{float}  
\usepackage{algorithm}
\usepackage{algorithmic}
\usepackage{amsmath}
\usepackage{import}
\usepackage{graphicx}
\usepackage{todonotes} % Hai added 
\usepackage[english]{babel} % Hai added 
\usepackage{amsthm} % Hai added
\usepackage{dsfont}
\usepackage{lipsum} % for filler text % Hai added 
\usepackage[mathscr]{euscript}
\newtheorem{thm}{Theorem}[section]

\usepackage[normalem]{ulem}
\usepackage{wrapfig,lipsum,booktabs}
\usepackage{subfig}
\usepackage{caption}
\newcommand{\norm}[1]{\left \| #1 \right \|}
\newcommand{\bpi}{{\pi}}
\newcommand{\bx}{\textbf{x}}
\newcommand{\bzero}{\textbf{0}}
\newcommand{\goto}{\rightarrow}
\newcommand{\simplex}{\Delta}

\title{Vision Transformer Visualization: What Neurons Tell and How Neurons Behave?}

\author{%
Van-Anh Nguyen\thanks{The first two authors contributed equally to this work} \\
  Monash University, Australia\\
  \texttt {van-anh.nguyen@monash.edu} \\
   \And
   Khanh Pham Dinh$^*$ \\
   OrientSoftware \\
  \texttt {khanh.pham@orientsoftware.com} \\
  \And
   Long Tung Vuong \\
  Monash University, Australia \\
  \texttt {longvt94@gmail.com} \\
  \And
   Thanh-Toan Do \\
  Monash University, Australia \\
  \texttt {toan.do@monash.edu} \\
  \And
   Quan Hung Tran \\
  Adobe Research, San Jose, CA, USA \\
  \texttt {qtran@adobe.com} \\
  \And
   Dinh Phung \\
  Monash University, Australia \\
  \texttt {dinh.phung@monash.edu} \\
  \And
   Trung Le \\
  Monash University, Australia \\
  \texttt {trunglm@monash.edu} \\
}

\begin{document}

\maketitle

\global\long\def\sidenote#1{\marginpar{\small\emph{{\color{Medium}#1}}}}%
\global\long\def\se{\hat{\text{se}}}%
\global\long\def\interior{\text{int}}%
\global\long\def\boundary{\text{bd}}%
\global\long\def\ML{\textsf{ML}}%
\global\long\def\GML{\mathsf{GML}}%
\global\long\def\HMM{\mathsf{HMM}}%
\global\long\def\support{\text{supp}}%
\global\long\def\new{\text{*}}%
\global\long\def\stir{\text{Stirl}}%
\global\long\def\mA{\mathcal{A}}%
\global\long\def\mB{\mathcal{B}}%
\global\long\def\expect{\mathbb{E}}%
\global\long\def\mF{\mathcal{F}}%
\global\long\def\mK{\mathcal{K}}%
\global\long\def\mH{\mathcal{H}}%
\global\long\def\mX{\mathcal{X}}%
\global\long\def\mZ{\mathcal{Z}}%
\global\long\def\mS{\mathcal{S}}%
\global\long\def\Ical{\mathcal{I}}%
\global\long\def\mT{\mathcal{T}}%
\global\long\def\Pcal{\mathcal{P}}%
\global\long\def\dist{d}%
\global\long\def\HX{\entro\left(X\right)}%
\global\long\def\entropyX{\HX}%
\global\long\def\HY{\entro\left(Y\right)}%
\global\long\def\entropyY{\HY}%
\global\long\def\HXY{\entro\left(X,Y\right)}%
\global\long\def\entropyXY{\HXY}%
\global\long\def\mutualXY{\mutual\left(X;Y\right)}%
\global\long\def\mutinfoXY{\mutualXY}%
\global\long\def\given{\mid}%
\global\long\def\gv{\given}%
\global\long\def\goto{\rightarrow}%
\global\long\def\asgoto{\stackrel{a.s.}{\longrightarrow}}%
\global\long\def\pgoto{\stackrel{p}{\longrightarrow}}%
\global\long\def\dgoto{\stackrel{d}{\longrightarrow}}%
\global\long\def\lik{\mathcal{L}}%
\global\long\def\logll{\mathit{l}}%
\global\long\def\bigcdot{\raisebox{-0.5ex}{\scalebox{1.5}{\ensuremath{\cdot}}}}%
\global\long\def\sig{\textrm{sig}}%
\global\long\def\likelihood{\mathcal{L}}%
\global\long\def\vectorize#1{\mathbf{#1}}%
\global\long\def\vt#1{\mathbf{#1}}%
\global\long\def\gvt#1{\boldsymbol{#1}}%
\global\long\def\idp{\ \bot\negthickspace\negthickspace\bot\ }%
\global\long\def\cdp{\idp}%
\global\long\def\das{}%
\global\long\def\id{\mathbb{I}}%
\global\long\def\idarg#1#2{\id\left\{  #1,#2\right\}  }%
\global\long\def\iid{\stackrel{\text{iid}}{\sim}}%
\global\long\def\bzero{\vt 0}%
\global\long\def\bone{\mathbf{1}}%
\global\long\def\a{\mathrm{a}}%
\global\long\def\ba{\mathbf{a}}%
\global\long\def\b{\mathrm{b}}%
\global\long\def\bb{\mathbf{b}}%
\global\long\def\B{\mathrm{B}}%
\global\long\def\boldm{\boldsymbol{m}}%
\global\long\def\c{\mathrm{c}}%
\global\long\def\C{\mathrm{C}}%
\global\long\def\d{\mathrm{d}}%
\global\long\def\D{\mathrm{D}}%
\global\long\def\N{\mathrm{N}}%
\global\long\def\h{\mathrm{h}}%
\global\long\def\H{\mathrm{H}}%
\global\long\def\bH{\mathbf{H}}%
\global\long\def\K{\mathrm{K}}%
\global\long\def\M{\mathrm{M}}%
\global\long\def\bff{\vt f}%
\global\long\def\bx{\mathbf{\mathbf{x}}}%
\global\long\def\bl{\boldsymbol{l}}%
\global\long\def\s{\mathrm{s}}%
\global\long\def\T{\mathrm{T}}%
\global\long\def\bu{\mathbf{u}}%
\global\long\def\v{\mathrm{v}}%
\global\long\def\bv{\mathbf{v}}%
\global\long\def\bo{\boldsymbol{o}}%
\global\long\def\bh{\mathbf{h}}%
\global\long\def\bs{\boldsymbol{s}}%
\global\long\def\x{\mathrm{x}}%
\global\long\def\bx{\mathbf{x}}%
\global\long\def\bz{\mathbf{z}}%
\global\long\def\hbz{\hat{\bz}}%
\global\long\def\z{\mathrm{z}}%
\global\long\def\y{\mathrm{y}}%
\global\long\def\bxnew{\boldsymbol{y}}%
\global\long\def\bX{\boldsymbol{X}}%
\global\long\def\tbx{\tilde{\bx}}%
\global\long\def\by{\mathbf{y}}%
\global\long\def\bY{\boldsymbol{Y}}%
\global\long\def\bZ{\boldsymbol{Z}}%
\global\long\def\bU{\boldsymbol{U}}%
\global\long\def\bn{\boldsymbol{n}}%
\global\long\def\bV{\boldsymbol{V}}%
\global\long\def\bI{\boldsymbol{I}}%
\global\long\def\J{\mathrm{J}}%
\global\long\def\bJ{\mathbf{J}}%
\global\long\def\w{\mathrm{w}}%
\global\long\def\bw{\vt w}%
\global\long\def\bW{\mathbf{W}}%
\global\long\def\balpha{\gvt{\alpha}}%
\global\long\def\bdelta{\boldsymbol{\delta}}%
\global\long\def\bsigma{\gvt{\sigma}}%
\global\long\def\bbeta{\gvt{\beta}}%
\global\long\def\bmu{\gvt{\mu}}%
\global\long\def\btheta{\boldsymbol{\theta}}%
\global\long\def\blambda{\boldsymbol{\lambda}}%
\global\long\def\bgamma{\boldsymbol{\gamma}}%
\global\long\def\bpsi{\boldsymbol{\psi}}%
\global\long\def\bphi{\boldsymbol{\phi}}%
\global\long\def\bpi{\boldsymbol{\pi}}%
\global\long\def\bomega{\boldsymbol{\omega}}%
\global\long\def\bepsilon{\boldsymbol{\epsilon}}%
\global\long\def\btau{\boldsymbol{\tau}}%
\global\long\def\bxi{\boldsymbol{\xi}}%
\global\long\def\realset{\mathbb{R}}%
\global\long\def\realn{\realset^{n}}%
\global\long\def\integerset{\mathbb{Z}}%
\global\long\def\natset{\integerset}%
\global\long\def\integer{\integerset}%
\global\long\def\natn{\natset^{n}}%
\global\long\def\rational{\mathbb{Q}}%
\global\long\def\rationaln{\rational^{n}}%
\global\long\def\complexset{\mathbb{C}}%
\global\long\def\comp{\complexset}%
\global\long\def\compl#1{#1^{\text{c}}}%
\global\long\def\and{\cap}%
\global\long\def\compn{\comp^{n}}%
\global\long\def\comb#1#2{\left({#1\atop #2}\right) }%
\global\long\def\param{\vt w}%
\global\long\def\Param{\Theta}%
\global\long\def\meanparam{\gvt{\mu}}%
\global\long\def\Meanparam{\mathcal{M}}%
\global\long\def\meanmap{\mathbf{m}}%
\global\long\def\logpart{A}%
\global\long\def\simplex{\Delta}%
\global\long\def\simplexn{\simplex^{n}}%
\global\long\def\dirproc{\text{DP}}%
\global\long\def\ggproc{\text{GG}}%
\global\long\def\DP{\text{DP}}%
\global\long\def\ndp{\text{nDP}}%
\global\long\def\hdp{\text{HDP}}%
\global\long\def\gempdf{\text{GEM}}%
\global\long\def\rfs{\text{RFS}}%
\global\long\def\bernrfs{\text{BernoulliRFS}}%
\global\long\def\poissrfs{\text{PoissonRFS}}%
\global\long\def\grad{\gradient}%
\global\long\def\gradient{\nabla}%
\global\long\def\partdev#1#2{\partialdev{#1}{#2}}%
\global\long\def\partialdev#1#2{\frac{\partial#1}{\partial#2}}%
\global\long\def\partddev#1#2{\partialdevdev{#1}{#2}}%
\global\long\def\partialdevdev#1#2{\frac{\partial^{2}#1}{\partial#2\partial#2^{\top}}}%
\global\long\def\closure{\text{cl}}%
\global\long\def\cpr#1#2{\Pr\left(#1\ |\ #2\right)}%
\global\long\def\var{\text{Var}}%
\global\long\def\Var#1{\text{Var}\left[#1\right]}%
\global\long\def\cov{\text{Cov}}%
\global\long\def\Cov#1{\cov\left[ #1 \right]}%
\global\long\def\COV#1#2{\underset{#2}{\cov}\left[ #1 \right]}%
\global\long\def\corr{\text{Corr}}%
\global\long\def\sst{\text{T}}%
\global\long\def\SST{\sst}%
\global\long\def\ess{\mathbb{E}}%
\global\long\def\Ess#1{\ess\left[#1\right]}%
% \newcommandx\ESS[2][usedefault, addprefix=\global, 1=]{\underset{#2}{\ess}\left[#1\right]}%
\global\long\def\fisher{\mathcal{F}}%
\global\long\def\bfield{\mathcal{B}}%
\global\long\def\borel{\mathcal{B}}%
\global\long\def\bernpdf{\text{Bernoulli}}%
\global\long\def\betapdf{\text{Beta}}%
\global\long\def\dirpdf{\text{Dir}}%
\global\long\def\gammapdf{\text{Gamma}}%
\global\long\def\gaussden#1#2{\text{Normal}\left(#1, #2 \right) }%
\global\long\def\gauss{\mathbf{N}}%
\global\long\def\gausspdf#1#2#3{\text{Normal}\left( #1 \lcabra{#2, #3}\right) }%
\global\long\def\multpdf{\text{Mult}}%
\global\long\def\poiss{\text{Pois}}%
\global\long\def\poissonpdf{\text{Poisson}}%
\global\long\def\pgpdf{\text{PG}}%
\global\long\def\wshpdf{\text{Wish}}%
\global\long\def\iwshpdf{\text{InvWish}}%
\global\long\def\nwpdf{\text{NW}}%
\global\long\def\niwpdf{\text{NIW}}%
\global\long\def\studentpdf{\text{Student}}%
\global\long\def\unipdf{\text{Uni}}%
\global\long\def\transp#1{\transpose{#1}}%
\global\long\def\transpose#1{#1^{\mathsf{T}}}%
\global\long\def\mgt{\succ}%
\global\long\def\mge{\succeq}%
\global\long\def\idenmat{\mathbf{I}}%
\global\long\def\trace{\mathrm{tr}}%
\global\long\def\argmax#1{\underset{_{#1}}{\text{argmax}} }%
\global\long\def\argmin#1{\underset{_{#1}}{\text{argmin}\ } }%
\global\long\def\diag{\text{diag}}%
\global\long\def\norm{}%
\global\long\def\spn{\text{span}}%
\global\long\def\vtspace{\mathcal{V}}%
\global\long\def\field{\mathcal{F}}%
\global\long\def\ffield{\mathcal{F}}%
\global\long\def\inner#1#2{\left\langle #1,#2\right\rangle }%
\global\long\def\iprod#1#2{\inner{#1}{#2}}%
\global\long\def\dprod#1#2{#1 \cdot#2}%
\global\long\def\norm#1{\left\Vert #1\right\Vert }%
\global\long\def\entro{\mathbb{H}}%
\global\long\def\entropy{\mathbb{H}}%
\global\long\def\Entro#1{\entro\left[#1\right]}%
\global\long\def\Entropy#1{\Entro{#1}}%
\global\long\def\mutinfo{\mathbb{I}}%
\global\long\def\relH{\mathit{D}}%
\global\long\def\reldiv#1#2{\relH\left(#1||#2\right)}%
\global\long\def\KL{KL}%
\global\long\def\KLdiv#1#2{\KL\left(#1\parallel#2\right)}%
\global\long\def\KLdivergence#1#2{\KL\left(#1\ \parallel\ #2\right)}%
\global\long\def\crossH{\mathcal{C}}%
\global\long\def\crossentropy{\mathcal{C}}%
\global\long\def\crossHxy#1#2{\crossentropy\left(#1\parallel#2\right)}%
\global\long\def\breg{\text{BD}}%
\global\long\def\lcabra#1{\left|#1\right.}%
\global\long\def\lbra#1{\lcabra{#1}}%
\global\long\def\rcabra#1{\left.#1\right|}%
\global\long\def\rbra#1{\rcabra{#1}}%

\begin{abstract}
%Vision Transformers (ViT) work like a charm for various tasks in computer vision. It is appealing to visualize the information from input images captured in neurons and feature embeddings across the ViT's layers. 
Recently vision transformers (ViT) have been applied successfully for various tasks in computer vision. However, important questions such as why they work or how they behave still remain largely unknown. 
In this paper, we propose an effective visualization technique, assisting us in exposing the information carried in neurons and feature embeddings across the ViT's layers. 
%Specifically, our visualization technique is developed based on the computational process of ViTs and supports us in visualizing the local and global information in input images represented by neurons and feature embeddings across the layers. 
Our approach departs from the computational process of ViTs with a focus on visualizing the local and global information in input images and the latent feature embeddings at multiple levels.
Visualizations at the input and embeddings at level 0 reveal interesting findings such as providing support as to why ViTs are rather generally robust to image occlusions and patch shuffling; or unlike CNNs, level 0 embeddings already carry rich semantic details.
%Furthermore, by observing and analyzing the outcome of our visualization technique, we harvest interesting findings of the filters in ViTs and the clustering behavior of feature embeddings relevant to object patches across layers. 
Next, we develop rigorous framework to perform effective visualizations across layers, exposing the effects of ViTs filters and grouping/clustering behaviors to object patches.
Finally, we provide comprehensive experiments on real datasets to qualitatively and quantitatively demonstrate the merit of our proposed methods as well as our findings. \url{https://github.com/byM1902/ViT_visualization}
%Finally, we visualize and conduct experiments on real images to qualitatively and quantitatively demonstrate our findings.
\end{abstract}

\section{Introduction}
Inspired by the success of Transformer \cite{vaswani2017attention}
in natural language processing \cite{devlin2018bert}, Vision Transformers
(ViT) \cite{dosovitskiy2021an,pmlr-v139-touvron21a} have been proposed
for vision data with a specific focus on large-scale and complex
datasets such as ImageNet \cite{deng2009imagenet,russakovsky2015imagenet}.
Subsequently, inherited from the original ViT \cite{dosovitskiy2021an},
many variants of ViTs have been explored to deal with various applications
in computer vision ranging from image classification \cite{pmlr-v139-touvron21a,chen2021crossvit},
object detection \cite{carion2020end,zhu2020deformable,liu2022dab},
semantic segmentation \cite{zheng2021rethinking,strudel2021segmenter,xie2021segformer},
image captioning \cite{cornia2020meshed,yu2019multimodal} to name
a few. 

With the engagement of the multi-head self-attention mechanism
\cite{vaswani2017attention}, ViTs operating principle is fundamentally
different from that of Convolutional Neural Networks (CNN) \cite{lecun1995convolutional,he2016deep}.
Therefore, it is essential to answer the questions regarding the intriguing
properties of ViTs. For instance, \textquotedbl\emph{how are ViTs
fundamentally different from CNNs?}\textquotedbl , \textquotedbl\emph{what
are the intriguing properties of ViTs?}\textquotedbl , or \textquotedbl\emph{how
are ViTs robust to adversarial attacks, image occlusions, or patch
shuffling?}\textquotedbl . Targeting these questions, some works have been dedicated
to studying the characteristics and behaviors of ViT, notably \cite{naseer2021intriguing,bhojanapalli2021understanding,shao2021adversarial,raghu2021vision,park2022vision}.

Particularly, Naseer et al. \cite{naseer2021intriguing} studied the
intriguing properties of ViTs and found that ViTs are more robust
to image occlusions, patch shuffling, and shape-biased learning than
CNNs. Shao et al. \cite{shao2021adversarial} analyzed ViTs against
adversarial noise and demonstrated that ViTs are more robust to high
frequency changes. Additionally, Bhojanapalli et al. \cite{bhojanapalli2021understanding}
studied ViTs against spatial perturbations \cite{shao2021adversarial}
and showed their robustness to the removal of almost any single layer
and their robustness superiority to the ResNet counterparts. Recently,
to compare the robustness of ViTs and CNNs, Yutong et al. \cite{bai2021transformers}
undertook comprehensive and fair experiments to come with a quite opposite
conclusion: ViTs are not actually more robust than CNNs w.r.t. adversarial
examples. Furthermore, Maithra et al. \cite{raghu2021vision} analyzed
the internal representation structure of ViTs and CNNs on image classification
benchmarks and found that ViTs have more uniform representations across
all layers. Namuk et al. \cite{park2022vision} studied the behavior
of multi-head self-attentions (MSA) and convolutional layers (Convs),
and then proposed to replace Conv blocks at the end of a stage %are replaced 
with MSA blocks for improving the performance. 

In this paper, to enrich the understanding of the behaviors and characteristics
of ViTs, we propose a visualization technique that allows us to visualize
the information in input images captured in neurons and feature embeddings
across the layers of ViTs. Furthermore, by observing and analyzing
the outputs of our visualization technique, we observe an interesting
clustering behavior of feature embeddings across the layers for which
we can develop a simplified mathematical explanation. More concretely,
our contributions in this paper can be summarized:
\begin{itemize}
\item We propose a visualization technique to visually expose the local/global
information in input images carried in neurons and embeddings across
the layers of ViTs. Additionally, using the proposed visualization
technique, we can partly explain why ViTs are rather robust to image
occlusions and patch shuffling.
\item Moreover, by visualizing neurons and embeddings at the layer 0, we
find that the filters of ViTs (i.e., $16\times16$ filters) are sufficiently
powerful to activate multiple object types across multiple images.
This is different from CNNs, where the low-level filters are typically
small and can capture only low-level features (e.g., edges or lines)
at low-level layers.
\item Furthermore, by analyzing the outputs of our visualization technique, we further realize
the clustering behavior of feature embeddings across layers. Specifically,
when moving up to higher layers, ViT's feature embeddings at each
layer tend to form more well-separated clusters, each of which is
dominated by feature embeddings corresponding to the same object type's
patches (e.g., feature embeddings of the dog patches);
the attention weights for feature embeddings in an object cluster
are also higher than others. Finally, we develop a simplified mathematical theory
to explain this clustering behavior and conduct experiments on the
Pascal VOC 2009 dataset \cite{Everingham10} to quantitatively verify
it. 
\end{itemize}

\section{Vision Transformers}
We recap the technical specification of ViT \cite{dosovitskiy2021an},
which lays foundation for the presentation of our visualization technique
in the sequel. Given an image $\mathbf{x}\in\mathbb{R}^{H\times W\times C}$,
we divide this image to $N$ patches $\left[\mathbf{x}_{p}^{i}\right]_{i=1}^{N}$,
each of which has the shape $\mathbf{x}_{p}^{i}\in\mathbb{R}^{P\times P\times C}$
with $N=\frac{H\times W}{P^{2}}$. We next apply $D$ filters with
the shape $P\times P\times C$ to each patch to work out a patch embedding
$\mathbf{\tilde{z}}_{0}^{i}\in\mathbb{R}^{D\times1}$ ($i=1,...,N)$.
We then concatenate the patch embedding with the class embedding $\mathbf{x}_{class}=\tilde{\mathbf{z}}_{0}^{0}$.
Finally, we add the position encoding $\mathbf{E}_{pos}\in\mathbb{R}^{(N+1)\times D}$
to obtain the patch embeddings and class embedding at the layer $0$
as\vspace{-2mm}
\[
\mathbf{z}_{0}=\left[\mathbf{x}_{class},\mathbf{\tilde{z}}_{0}^{1},...,\mathbf{\tilde{z}}_{0}^{N}\right]^{T}+\mathbf{E}_{pos}\in\mathbb{R}^{(N+1)\times D}.
\]
\vspace{-2mm}

The representations at layer $0$ are fed to the Transformer encoder
\cite{transformer} consisting of alternating layers of multi-headed
self-attention (MSA) and MLP blocks. Layer norm (LN) is applied before
every block, and residual connections after every block. Additionally,
the MLP contains two layers with a GELU non-linearity.\vspace{-2mm}

\begin{alignat*}{1}
\tilde{\mathbf{z}}_{l} & =MSA\left(LN\left(\mathbf{z}_{l-1}\right)\right)+\mathbf{z}_{l-1},\,\,\,\,l=1,...,L\\
\mathbf{z}_{l} & =MLP\left(LN\left(\tilde{\mathbf{z}}_{l}\right)\right)+\tilde{\mathbf{z}}_{l},\,\,\,\,\,\,\,\,\,l=1,...,L\\
\mathbf{y} & =LN\left(\mathbf{z}_{L}^{0}\right).
\end{alignat*}

Here we note that we use the pretrained ViT \cite{dosovitskiy2021an}
for the visualization and empirical studies of the behaviors of ViT.
Therefore, we have $H=W=224$, $D=768$, and $N=14\times14=196$.

\section{Visualization for ViT}
In this section, we first present our visualization technique for neurons and embeddings across the layer 0 and subsequent layers. We then discuss the findings based on the observation of the visual outputs. Interestingly, the visual outputs give us a hint on the clustering behavior of the patch embeddings corresponding to the objects in an input image. We finally conduct further experiments on real images to quantitatively confirm this finding. 

\subsection{What Do Neurons Tell?}
\subsubsection{Visualization for Layer 0}

We visualize to answer the question: ``\emph{What information is
captured or represented by a neuron or an embedding at the layer $0$?}''.
Since we convolve a \emph{specific filter} with \emph{an image patch}
to compute a neuron at the layer $0$ and \emph{all $D$ filters}
with \emph{an image patch} to form a patch embedding, a neuron or
patch embedding at the layer $0$ can only capture a local information
of a patch. 

\textbf{Visualizing a neuron and an embedding at the layer $\mathbf{0}$:}
The neuron $\mathbf{z}_{0}^{i,j}$ (i.e., the neuron $j$ of the patch
embedding $i$) is formed by convolving the filter $\mathbf{f}_{j}\in\mathbb{R}^{P\times P\times C}$
($1\leq j\leq D$) and the patch $\mathbf{x}_{p}^{i}$. Therefore,
this neuron contains a local information of the patch $i$ w.r.t.
the filter $\mathbf{f}_{j}$, meaning that it captures local information
from a view of a patch. To visualize the neuron $\mathbf{z}_{0}^{i,j}$,
we perform the element-wise product between the patch and the filter:
$\mathcal{V}(\mathbf{z}_{0}^{i,j})=\mathbf{f}_{j}\otimes\mathbf{x}_{p}^{i}\in\mathbb{R}^{P\times P\times C}$.
Here we denote $\mathcal{V}(\mathbf{z}_{0}^{i,j})$ as
the \emph{visualization of the neuron} $\mathbf{z}_{0}^{i,j}$. Furthermore,
to visualize the patch embedding $\mathbf{z}_{0}^{i}$ (i.e., the
row $\mathbf{z}_{0}^{i,1:D}$), we overlay the visualizations $\mathcal{V}(\mathbf{z}_{0}^{i,1:D})$.
In other words, we stack the $D$ tensors/images $\mathcal{V}(\mathbf{z}_{0}^{i,j})\in\mathbb{R}^{P\times P\times C}$
to represent the information of the patch embedding $\mathbf{z}_{0}^{i}$.
Obviously, the visualization $\mathcal{R}(\mathbf{z}_{0}^{i})$
of $\mathbf{z}_{0}^{i}$ can only capture a local information according
to a multi-view of the patch $\mathbf{x}_{p}^{i}$.

\begin{figure}
\begin{centering}
\includegraphics[width=1\textwidth]{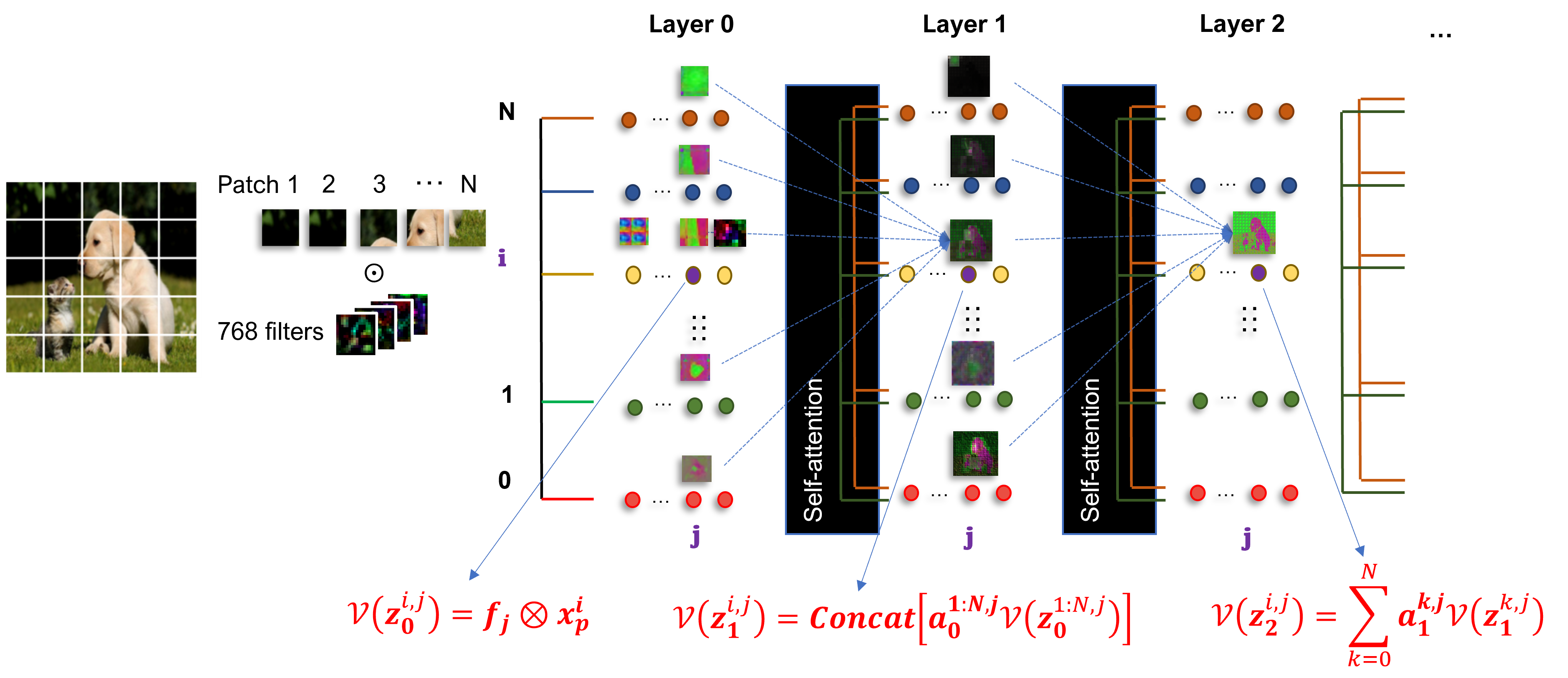}\vspace{-2mm}
\par\end{centering}
\caption{Overview of our visualization technique. At the layer $0$, the visualization
% of the neuron $\mathbf{z}_{0}^{i,j}$ is $\mathcal{V}\left(\mathbf{z}_{0}^{i,j}\right)=\mathbf{f}_{j} \varotimes \mathbf{x}_{p}^{i} \in \mathbb{R}^{P \times P \times C}$.
of the neuron $\mathbf{z}_{0}^{i,j}$ is $\mathcal{V}\left(\mathbf{z}_{0}^{i,j}\right)=\mathbf{f}_{j} \bigotimes \mathbf{x}_{p}^{i} \in \mathbb{R}^{P \times P \times C}$.
At the layer $1$, we arrange $\mathbf{a}_{0}^{k,i}\mathcal{\mathcal{V}}\left(\mathbf{z}_{0}^{k,j}\right)\in\mathbb{R}^{P\times P\times C},k=1,...,N$
to form an image for $\mathcal{\mathcal{V}}\left(\mathbf{z}_{0}^{k,j}\right)=\text{Concat}\left[\mathbf{a}_{0}^{1:N,i}\mathcal{\mathcal{V}}\left(\mathbf{z}_{0}^{1:N,j}\right)\right]$.
At the layer $2$, we employ $\mathcal{V}(\mathbf{z}_{2}^{i,j})=\sum_{k=0}^{N}\mathbf{a}_{1}^{k,i}\mathcal{V}(\mathbf{z}_{1}^{k,j})$.
\label{fig:overview}}
\vspace{-6mm}
\end{figure}

Moreover, we visualize all neurons in the column $j$ corresponding
to the filter $\mathbf{f}_{j}$ (i.e., the column $\mathbf{z}_{0}^{1:N,j}$)
by placing $\mathcal{\mathcal{V}}(\mathbf{z}_{0}^{1:N,j})$
in the shape of input image. This can be realized by simply moving
the filter $\mathbf{f}_{j}$ across all patches and doing the element-wise
products on the fly. Certainly, the visualization $\mathcal{R}(\mathbf{z}_{0}^{1:N,j})$
can capture a global information of an entire image w.r.t. a view or
a filter. Our visualization for the layer $0$ is summarized in Figure \ref{fig:overview}.

\begin{figure}
\begin{centering}
\includegraphics[width=1\textwidth]{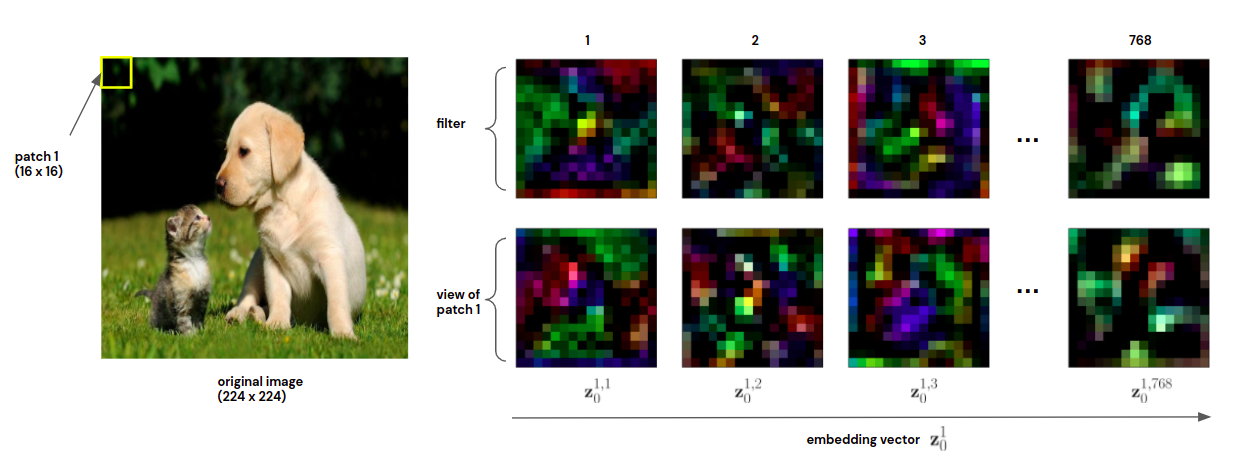}\vspace{-2mm}
\caption{Visualization of the neurons and the patch embedding $1$ on the layer
$0$.\label{fig:patch_embedd_lay0}}
\vspace{-4mm}
\par\end{centering}
\end{figure}

\textbf{\emph{Visualization of a specific neuron and an embedding
on the layer 0:}} We now visualize \emph{the neurons on the patch
embedding 1} in Figure \ref{fig:patch_embedd_lay0}. Each neuron contains
a local information of the image patch 1 or a view of this image patch.
Moreover, we overlay the visualizations of $768$ neurons in the row
1 together to yield the visualization of the $768$-dimensional patch
embedding.

\begin{figure}
\centering{}\vspace{-2mm}
\includegraphics[width=1\textwidth]{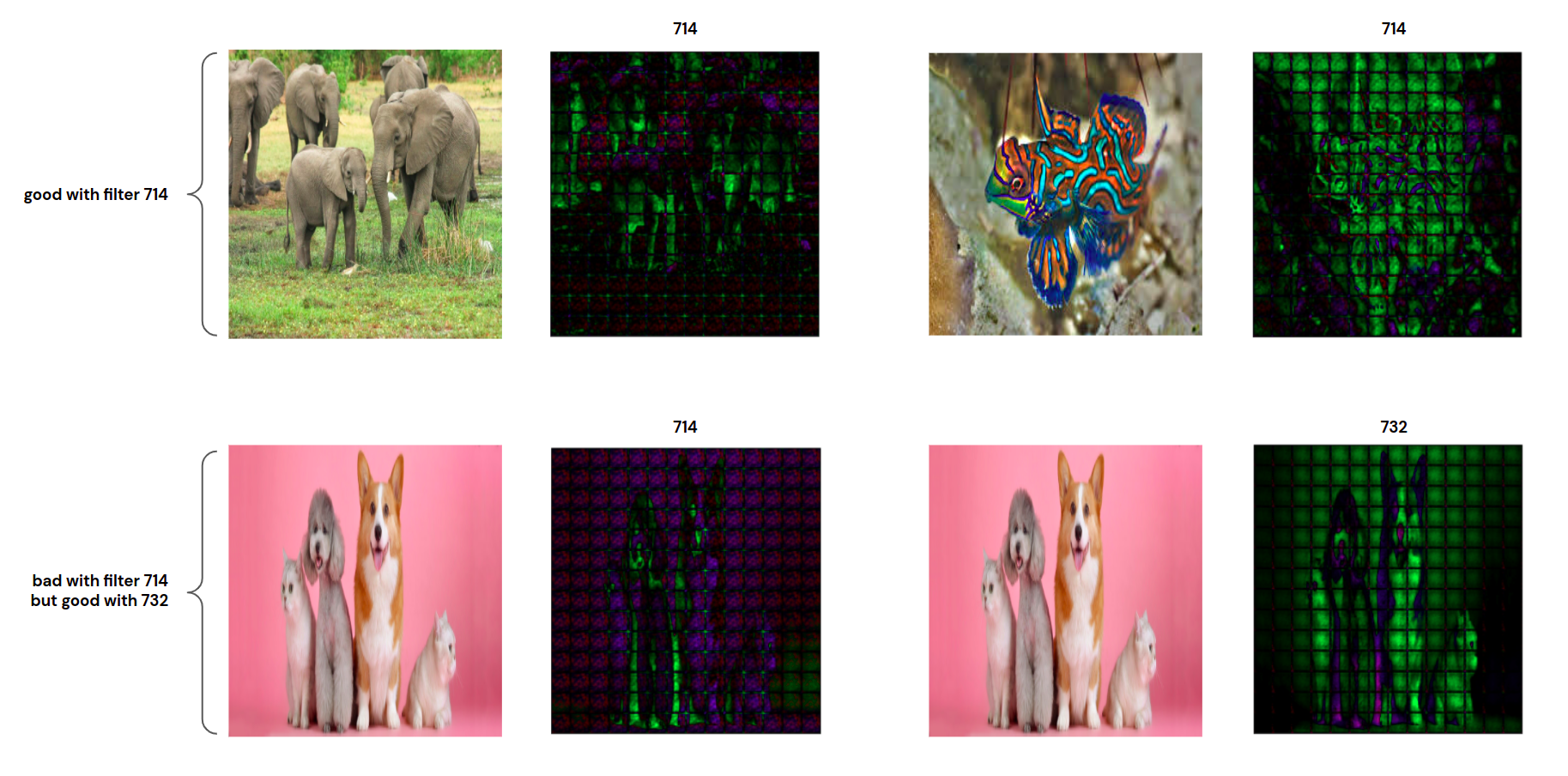}\vspace{-2mm}
\caption{Visualization of the neurons in the columns $\mathbf{z}_{0}^{1:N,714}$
corresponding to the filter $714$ and $\mathbf{z}_{0}^{1:N,732}$
corresponding to the filter $732$ for some input images. The filter
$714$ is good to activate elephants and fishes, whereas the filter $732$
is good to activate dogs and cats. \label{fig:lay0_good_bad_filters}}
\vspace{-4mm}
\end{figure}

\textbf{\emph{Visualization of a column of neurons corresponding to
a filter on the layer 0:}} We choose some filters and visualize the
neurons corresponding to these filters on the layer 0 in Figure \ref{fig:lay0_good_bad_filters}.
The neurons corresponding to a filter can be represented by an image
formed by convolving this filter and the input image. In other words,
these neurons represent a single-view of the entire image. We observe
that a $16\times16\times3$ filter of ViT is sufficiently powerful, hence it can
be learned to optimize and emphasize many object types across various
images. For instance, the filter $714$ can activate various elephants
with different sizes, fishes, and etc. This makes sense because the
large $16\times16\times3$ filters of ViT make a strong optimization
problem with $16\times16\times3$ variables to optimize on various
object types across images. It is worth noting that for state-of-the-art
CNNs \cite{he2016deep,tan2019efficientnet}, much smaller $3\times3\times3$
or $5\times5\times3$ filters are employed, hence the low-level filters
can only learn low-level features such as vertical/horizontal lines
or edges, whereas the higher-level filters can combine low-level features
to obtain more abstract features. We further observe that the filters
are complementary in the sense that each of them is responsible for
a set of object types. For instance, although the filter $714$ cannot
activate the dog or cat objects, the filter $732$ can complement
it to activate these object types. 

\subsubsection{Visualization for Subsequent Layers}

\textbf{Visualization for Layer 1:} We answer the same question but
for the neurons and embeddings on the subsequent layers. We start
with the layer 1. The neuron $\mathbf{z}_{1}^{i,j}$ is computed based
on the neurons $\mathbf{z}_{0}^{0:N,j}$ on the previous layer. Each
$\mathbf{z}_{0}^{i,j}$ represents the view $j$ of the patch $i$,
hence the neuron $\mathbf{z}_{1}^{i,j}$ combining $\mathbf{z}_{0}^{0:N,j}$
contains a global information of an entire image. Our purpose
is to visualize the information in an input image captured in a neuron
at a layer. Therefore, for the sake of visualization, we mainly
focus on the multi-head self-attention and consider how to visualize
the information of $\mathcal{V}\left(\mathbf{z}_{1}^{i,j}\right)$
based on $\mathcal{\mathcal{V}}\left(\mathbf{z}_{0}^{0:N,j}\right)$
and the attention weights $\mathbf{a}_{0}^{0:N,j}$, where $\mathbf{a}_{0}^{k,j}$
represents the attention weight of the embeddings $\mathbf{z}_{0}^{k}$
and $\mathbf{z}_{0}^{j}$. Visualization at this layer is summarized
in Figure \ref{fig:overview}.

Because the neuron $\mathbf{z}_{1}^{i,j}$ contains a global information
of an entire image and computationally depends on $\mathbf{z}_{0}^{0:N,j}$
and the attention weights $\mathbf{a}_{0}^{0:N,j}$ via a self-attention,
we visualize $\mathcal{V}\left(\mathbf{z}_{1}^{i,j}\right)$ by arranging
$\mathbf{a}_{0}^{k,i}\mathcal{\mathcal{V}}\left(\mathbf{z}_{0}^{k,j}\right)\in\mathbb{R}^{P\times P\times C},k=1,...,N$
to form an image or a tensor $\mathcal{V}\left(\mathbf{z}_{1}^{i,j}\right)\in\mathbb{R}^{H\times W\times C}$
in the order according to $k$. By doing so, we can summarize the
information of $\mathcal{\mathcal{V}}\left(\mathbf{z}_{0}^{k,j}\right)$
with the attention level $\mathbf{a}_{0}^{k,j}$ for $k=1,...,N$
into $\mathcal{V}\left(\mathbf{z}_{1}^{i,j}\right)$. Additionally,
to take into account multiple heads, we use the average of the attention
weights across the heads for our visualization. Moreover, we do
not use the information from $\mathbf{z}_{0}^{0,j}$ in the class
embedding when visualizing $\mathbf{z}_{1}^{i,j}$ since it does not
relate directly to a patch. Finally, to yield the visualization $\mathcal{V}\left(\mathbf{z}_{1}^{i}\right)=\mathcal{V}\left(\mathbf{z}_{1}^{i,1:D}\right)$
for the embedding $i$ at the layer $1$, we simply stack $\mathcal{V}\left(\mathbf{z}_{1}^{i,1:D}\right)$.

\begin{figure}[h]
\begin{centering}
\vspace{-2mm}
\includegraphics[width=1\textwidth]{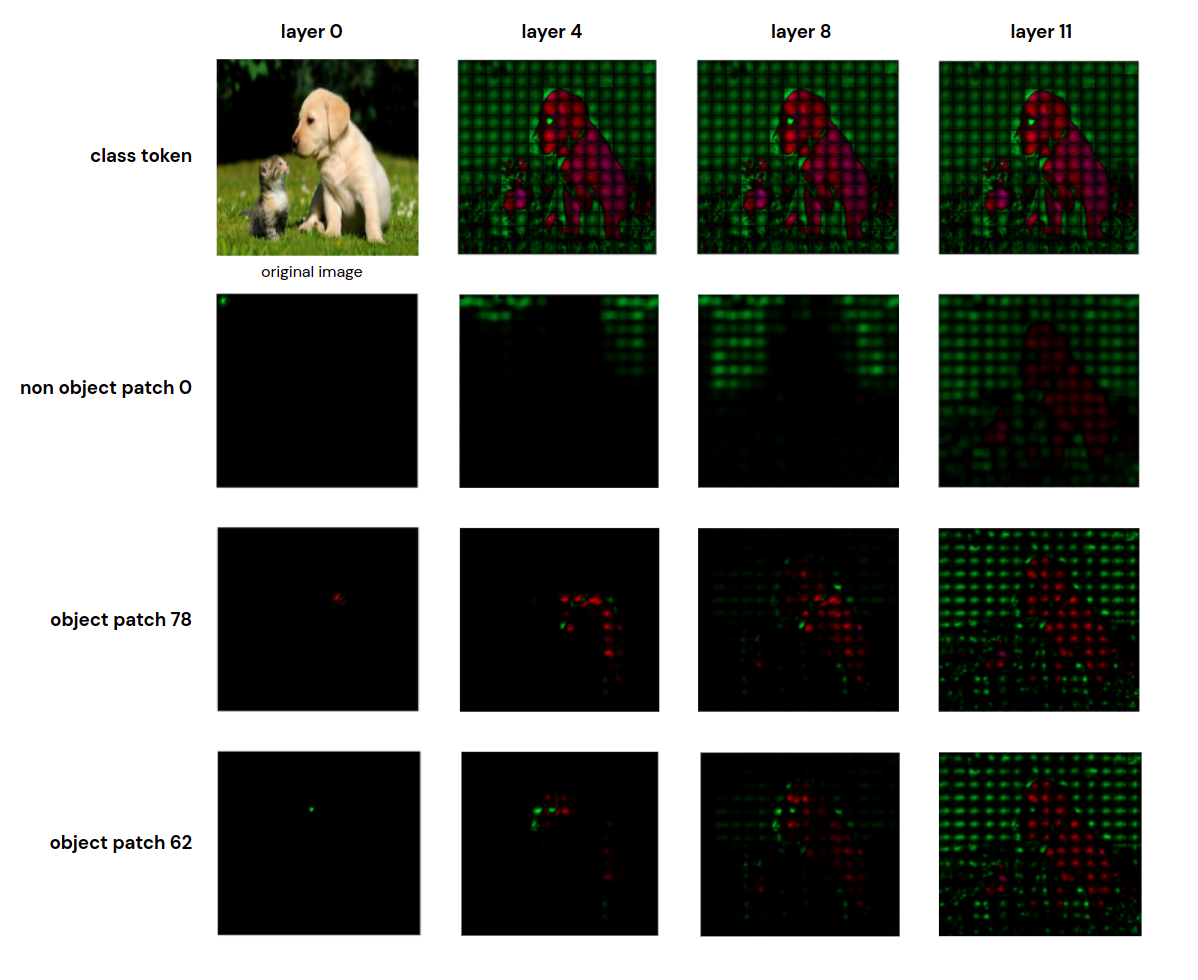}
\par\end{centering}
\begin{centering}
\vspace{-2mm}
\caption{Visualization of the neurons corresponding to background (none-object)
and foreground (object) patches.\label{fig:neurons_many_layers}}
\par\end{centering}
\centering{}\vspace{-2mm}
\end{figure}

\textbf{Visualization for Layer $\boldsymbol{\geq2}$:} We present
how to visualize the neurons and embeddings on the layer $2$ based
on those on the layer $1$. We apply the same way to visualize for
the remaining layers. Recap that each neuron $\mathbf{z}_{1}^{i,j}$
on the layer $1$ contains a global information of an entire image.
Because $\mathbf{z}_{2}^{i,j}$ computationally depends on the neurons
$\mathbf{z}_{1}^{0:N,j}$ with the attention weights $\mathbf{a}_{1}^{0:N,j}$,
we obtain the visualization $\mathcal{V}\left(\mathbf{z}_{2}^{i,j}\right)$
for the neuron $\mathbf{z}_{2}^{i,j}$ by $\mathcal{V}\left(\mathbf{z}_{2}^{i,j}\right)=\sum_{k=0}^{N}\mathbf{a}_{1}^{k,i}\mathcal{V}\left(\mathbf{z}_{1}^{k,j}\right),$
which returns an image or a tensor with the shape $\mathbb{R}^{H\times W\times C}$,
meaning that the neuron $\mathbf{z}_{2}^{i,j}$ contains a global
information of an entire image. Visualization at this layer is summarized
in Figure \ref{fig:overview}.

\textbf{\emph{Visualization of neurons and embeddings across layers:}}
In Figure \ref{fig:neurons_many_layers}, we visualize the neurons
corresponding to the filter $3$ across the layers for some cases:
(i) the neuron relevant to the class embedding, (ii) the neuron relevant
to a background (non-object) patch, and (iii) the neuron relevant
to a foreground (object) patch. We have the following observations
in order. \emph{First}, the neuron on the class embedding at high
layers contain sufficient information of the objects for making predictions.
This is possibly due to the fact that we make predictions on the class
embedding directly. \emph{Second}, the neurons corresponding to the
patches 62 and 78 (i.e., two patches on the dog object) contain a
clear information of the dog object, which is strengthened at higher
layers. \emph{Third}, the neuron on the background patch also contains
the information of the objects due to the self-attentions, but the
information amount are less than the neurons on the foreground patches.
\emph{Fourth}, the positional encoding information is retained in
the neurons. \emph{Fifth}, the visualizations of the neurons belonging
to the same object (e.g., the patches 62 and 78 for the dog object)
are more similar to those in other objects or background. This hints
us about the \emph{clustering behavior of the embeddings} corresponding
to the foreground patches of the same object, leading to our further
investigation in Section \ref{subsec:neuron_behave}. 

\textbf{\emph{Visualization of the information retained with an occlusion:}}
Inspired by \cite{naseer2021intriguing}, we visualize the information
from an input image retained in a neuron at a certain layer with an
occlusion. We examine three cases: (i) \emph{Random drop}, (ii) \emph{Non-salient
drop}, and (iii) \emph{Salient drop}. For the random drop, because
the patches are randomly dropped, the object information is partly
retained rather well. Additionally, the information from the dropped
patches seems hard to be recovered in the higher layers, which hurts
the final prediction. For the salient drop, the object information
retained a little, whereas for the non-salient drop, the object information
is remained rather intact. These concur with the observations in \cite{naseer2021intriguing}.

\begin{figure}
\centering{}\vspace{-2mm}
\includegraphics[width=1\textwidth]{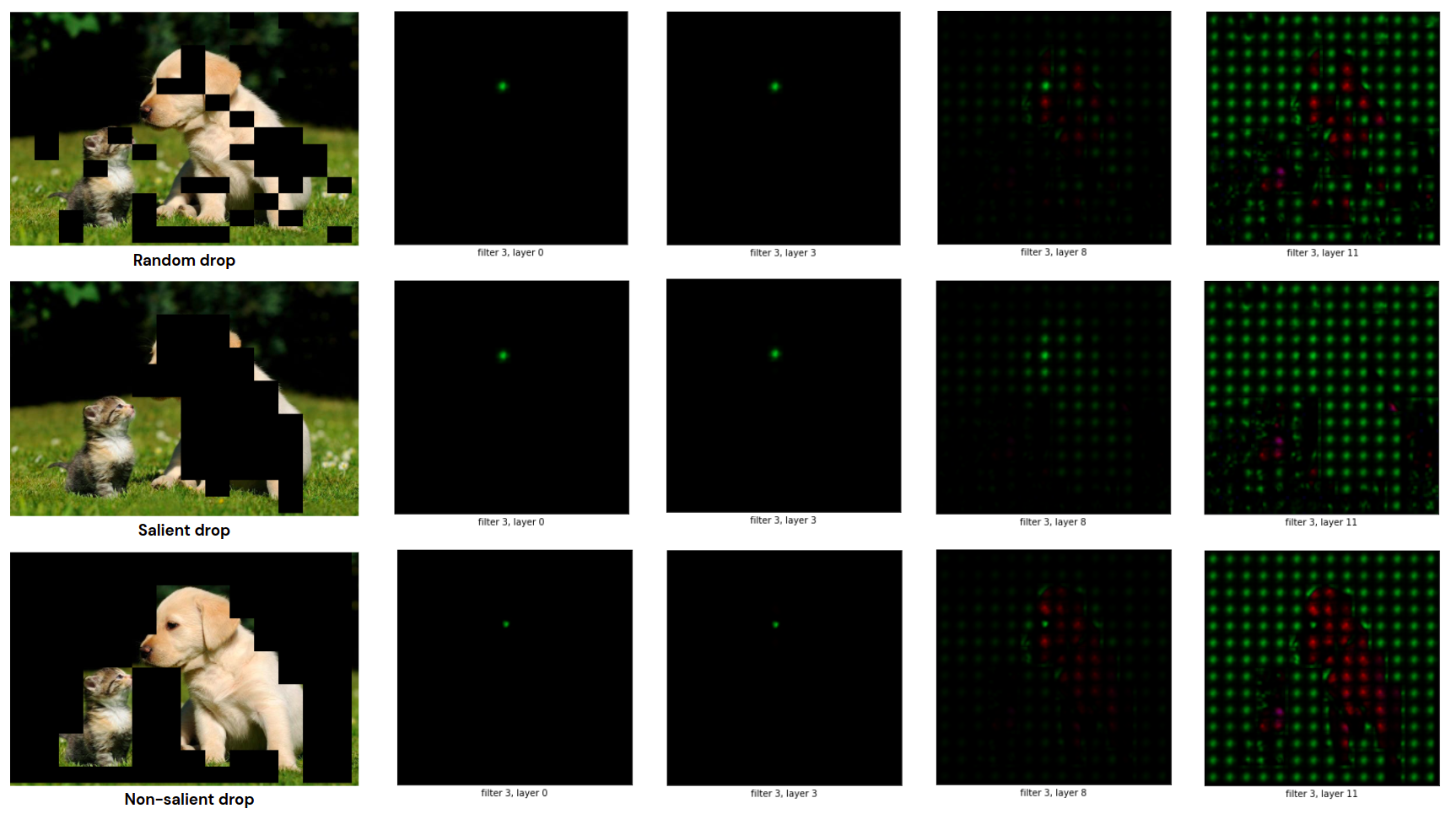}\vspace{-2mm}
\caption{Visualization of the information retained with an occlusion. Information
retained a little with the salient drop, retained more with the random
drop, and remained rather intact with the non-salient drop. \label{fig:all_drops}}
\vspace{-4mm}
\end{figure}

\subsection{How Do Neurons Behave?}
\label{subsec:neuron_behave}
As hinted by our clustering behavior conjecture in the previous section,
we quantitatively and qualitatively investigate this behavior on real
images. To realize this clustering behavior, we apply t-SNE \cite{JMLR:v9:vandermaaten08a}
to visualize the high-dimensional patch embeddings across the layers. 

\begin{figure}
\vspace{-2mm}
\subfloat[t-SNE for a single image with multiple objects.\label{fig:t-SNE_single_img}]{\begin{centering}
\includegraphics[width=0.98\textwidth]{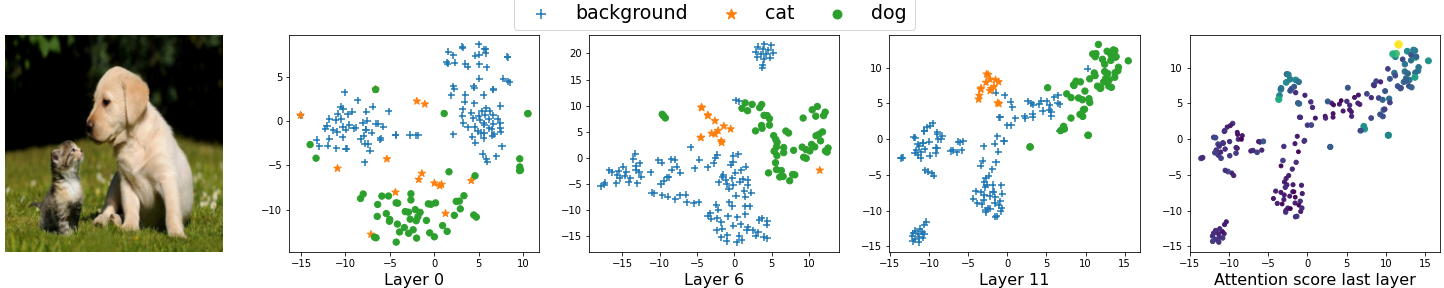}
\par\end{centering}

}\vspace{-2mm}

\subfloat[t-SNE for the concatenation of four images.\label{fig:t-SNE_multi_img} ]{\begin{centering}
\includegraphics[width=0.98\textwidth]{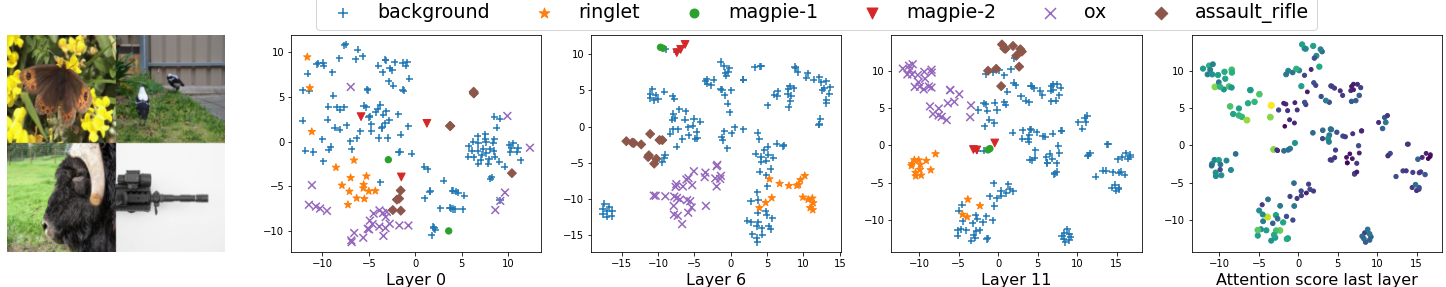}
\par\end{centering}

}\vspace{-2mm}

\caption{t-SNE of feature embeddings across the layers of ViT. At higher layers,
the object-patch embeddings form more separated clusters. Meanwhile,
the attention weights w.r.t. the class token of object-patch embeddings
are higher than others. \label{fig:t-SNE_embeddings}}

\centering{}\vspace{-6mm}
\end{figure}

\textbf{t-SNE visualization of the embeddings across the layers:}
We use t-SNE \cite{JMLR:v9:vandermaaten08a} to visualize the patch
embeddings at each layer of ViT for input images in Figure \ref{fig:t-SNE_embeddings}.
Particularly, in Figure \ref{fig:t-SNE_single_img}, we visualize
the patch embeddings for a single image with a dog object and a cat
object at the layers 0, 6, and 11. It can be observed that when
moving up to higher layers, the \emph{cluster for the dog patch embeddings}
(i.e., the green points) becomes more separated with the \emph{cluster
for the cat patch embeddings }(i.e., the orange points). In Figure
\ref{fig:t-SNE_multi_img}, we concatenate four images to have an
input image with diverge objects and then visualize the embeddings
at the layers 0, 6, and 11. Again, we observe the same clustering
behavior, i.e., the clusters for the object embeddings become more
separated at higher layers. In addition, at the last layer, we visualize
the \emph{attention weights w.r.t. the class token} for patch embeddings
with a bigger size and a brighter color for a higher value. It can
be seen that the object-patch embeddings have higher attention weights

In what follows, we provide a sketch mathematical explanation for
the clustering behavior. According to our analysis, this exciting
behavior mainly comes from the nature of the self-attention mechanism,
which makes two similar embeddings to become even more similar. To
simplify the context, we only consider the single-head self-attention
case.

\textbf{Mathematical explanation for the clustering behavior:} We
consider the single-head self-attention with the key matrix $W_{K}$,
the query matrix $W_{Q}$, and the value matrix $W_{V}$. We examine
the embeddings at two consecutive layers: $\mathbf{z}_{t}\in\mathbb{R}^{(N+1)\times D}$
and $\mathbf{z}_{t+1}\in\mathbb{R}^{(N+1)\times D}$, where each row
of them is an embedding. The attention weights $\mathbf{a}_{t}^{k,j}$
are computed based on the similarity of the linear projections of
$\mathbf{z}_{t}^{k}$ and $\mathbf{z}_{t}^{j}$ w.r.t. $W_{K}$ and
$W_{Q}$ respectively. We then apply the linear projection to $\mathbf{z}_{t}^{j},j=0,...,N$,
then compute $\tilde{\mathbf{z}}_{t+1}^{j},j=0,...,N$, and finally
do another linear projection to project $\tilde{\mathbf{z}}_{t+1}^{j},j=0,...,N$
back to the embedding space to obtain $\mathbf{z}_{t+1}^{j},j=0,...,N$.
To simplify our analysis, we further assume that $W_{V}=\mathbb{I}$
(i.e., the identity matrix). Therefore, we compute the embeddings
in the next layer as $\mathbf{z}_{t+1}^{j}=\sum_{k=0}^{N}\mathbf{a}_{t}^{k,j}\mathbf{z}_{t}^{k}$.
We develop the following theorem regarding the clustering behavior
of the embeddings.
\begin{thm}
\label{thm:intra_distance}(Proof can be found in the supplementary material) Let $\text{d}\left(A\right)=\max_{\mathbf{a},\mathbf{b}\in A}\Vert\mathbf{a}-\mathbf{b}\Vert_{2}$
be the diameter of the set $A$. The following statements hold

i) $\text{d\ensuremath{\left(Z_{t+1}\right)}\ensuremath{\ensuremath{\leq\text{d}\left(Z_{t}\right)}}}$,
where $Z_{t}$ and $Z_{t+1}$ are the sets of the row vectors (i.e.,
the embeddings) of $\mathbf{z}_{t}$ and $\mathbf{z}_{t+1}$ respectively.

ii) Assume that the embeddings $Z_{t}$ (or $\mathbf{z}_{t}$) can
be grouped into $M$ clusters $Z_{t}^{m}=\left\{ \mathbf{z}_{t}^{j}:j\in G^{m}\right\} $,
where $\left[G^{m}\right]_{m=1}^{M}$ is a partition of $\left\{ 0,1,...,N\right\} $
such that if $j\in G^{m}$ and $j'\in G^{m'}$ with $m\neq m'$, the
attention weight $\epsilon_{l}<\mathbf{a}_{t}^{j',j}<\epsilon_{u}$
for $\epsilon_{u}\geq\epsilon_{l}\geq0$. This condition means that
if $\mathbf{z}_{t}^{j}$ and $\mathbf{z}_{t}^{j'}$ belong to two
different clusters, they are less similar and their attention weight
$\mathbf{a}_{t}^{j',j}$ is possibly small. Let $N_{m}=\left|G_{m}\right|$
be the cardinality of the cluster $m$. For any $m$, we have the
following inequality:
\begin{align}
\text{d}\left(Z_{t+1}^{m}\right) & <\left(1-A_{m}\epsilon_{l}\right)^{2}\text{d}\left(Z_{t}^{m}\right)+A_{m}\epsilon_{u}\text{d}\left(Z_{t}\right)\left[A_{m}\epsilon_{u}\text{d}\left(Z_{t}\right)+2\right],\label{eq:distance_reduce}
\end{align}
where $A_m= N+1-N_m$ and $Z_{t+1}^{m}=\left\{ \mathbf{z}_{t+1}^{i}:i\in G^{m}\right\} $
is the cluster in the next layer. Note that $\text{d}\left(Z_{t+1}^{m}\right)$
and $\text{d}\left(Z_{t}^{m}\right)$ are exactly the intra-cluster
distances for these clusters.
\end{thm}
Theorem \ref{thm:intra_distance} asymptotically explains the clustering
behavior. First, Theorem \ref{thm:intra_distance}i discloses that
the diameter of the set of the next embeddings gets smaller. Second,
Theorem \ref{thm:intra_distance}ii states that if the clusters become
gradually well-separated, $\epsilon_{u}\geq\epsilon_{l}$ gets smaller
and approaches $0$, asymptotically meaning that $\text{d}\left(Z_{t+1}^{m}\right)\leq\text{d}\left(Z_{t}^{m}\right)$,
hence the intra-distance of a cluster gets decreased at the next layer.
We notice that because the object patches of the same object are possibly
similar, their embeddings at the layer $0$ are more similar. Therefore,
we possibly depart with quite good clusters. However, the difference
among embeddings are tiny due to initial values for the weight matrices.
We next experiment on real images to quantitatively verify the clustering
behavior.

\begin{wrapfigure}{o}{0.35\columnwidth}%
\begin{centering}
\includegraphics[width=0.3\textwidth]{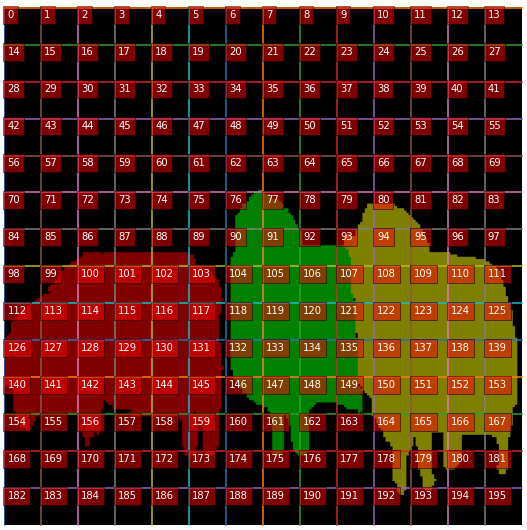}\vspace{-1mm}
\caption{How to label patches with object type labels or background.}
\par\end{centering}
\vspace{-4mm}
\end{wrapfigure}%

\textbf{\emph{Quantitative verification of the clustering behavior:}}\emph{
}We experiment on the real images from the Pascal VOC 2009 dataset
\cite{Everingham10}. %for the semantic segmentation task. 
We select
the 500 images and use the ground-truth pixel label to label an object
type or a background for a image patch. Additionally, we notice that
objects in the ImageNet dataset often take a big area in images, hence
to assign an object label for an image patch, this patch needs to
be overlapped at least 40\% with the object. Using this criterion,
we select images with at least two objects, wherein each object has
at least 3 patches validly belonging to it.

For each selected image, we feed to the ViT to extract the embeddings
at the layers. We employ the DBSCAN clustering algorithm \cite{martin2001database}
to cluster the embeddings at each layer. For the clustering solution
at each layer, we measure the purity score \cite{pml1Book}, the Silhouette
score \cite{silhouette}, the average cosine similarity of the embeddings
in a cluster, the average cosine similarity of the embeddings relevant
to an object, and the unique label ratio (cf. the supplementary material
for the formula of the clustering measures). 

\begin{figure}[h]
\begin{centering}
\vspace{-2mm}
\includegraphics[width=0.40\textwidth]{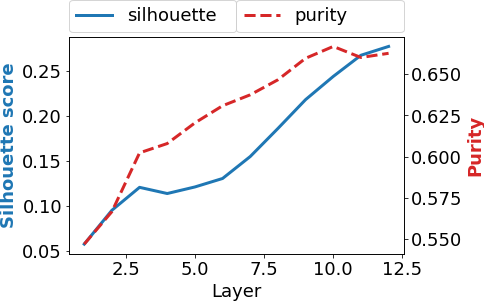}\hspace{5mm}\includegraphics[width=0.40\textwidth]{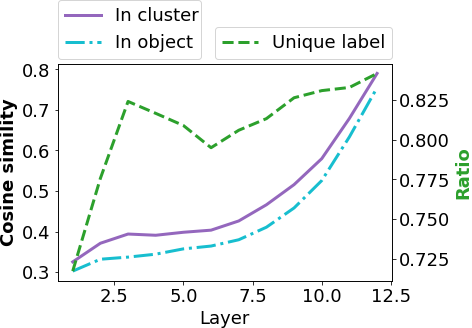}\vspace{-2mm}
\par\end{centering}
\caption{The visualization of the average the clusteting measures on the Pascal
VOC dataset across the ViT's layers. Left: the purity and Silhouette
scores. Right: the average cosine similarity of the embeddings in
a cluster, the average cosine similarity in of the embeddings relevant
to an object, and the unique label ratio. At higher layers, we obtain
more pure object-dominated clusters with high similar object-patch
embeddings. \label{fig:clustering_measures}\vspace{-3mm}
}
\end{figure}

A\emph{ high purity} means that each cluster is pure, meaning that
there exists a particular object or background so that its patches
really dominate this cluster. A \emph{high Silhouette} score implies
the intra-cluster distances are relatively smaller than the inter-cluster
distances. A\emph{ high average cosine similarity of the embeddings}
in a cluster means that the embeddings in a cluster are highly similar,
whereas a \emph{high average cosine similarity of the embeddings relevant
to an object }means that the embeddings corresponding to object patches
are highly similar. Finally, the unique label ratio is computed as
the ratio of the number of unique cluster labels (i.e., the cluster
label is the object or background patches dominating the cluster)
and the number of labels (i.e., the number of objects plus 1), hence
a \emph{high unique label rate} implies that %there exists a cluster for most of objects.
 there are more object-dominated clusters.

Figure \ref{fig:clustering_measures} visualizes the clustering measures
across the layers on the Pascal VOC dataset. It can be observed that
the purity becomes increased at higher layers, indicating purer clusters.
The Silhouette score gets increased, indicating the intra-cluster
distances decrease faster than the inter-cluster distances when moving
to higher layers. The average cosine similarity of the embeddings
in a cluster gets increased, indicating the embeddings in a cluster
becomes more similar at higher layers. The average cosine similarity
of the embeddings relevant to an object gets increased, indicating
the embeddings of object patches becomes more similar at higher layers.
The unique label ratio gets increased, indicating more object-dominated
clusters at higher layers. Putting altogether, at higher layers, we
obtain more pure object-dominated clusters with high similar object-patch
embeddings, concurring with the clustering behavior.

\begin{figure}[h]
\vspace{-2mm}
\subfloat[Image shuffled with the grid $2\times2$. The object embeddings form
clusters and attention weights for object embeddings are high.\label{fig:t-SNE_shuffle_2}]{\begin{centering}
\includegraphics[width=0.95\textwidth]{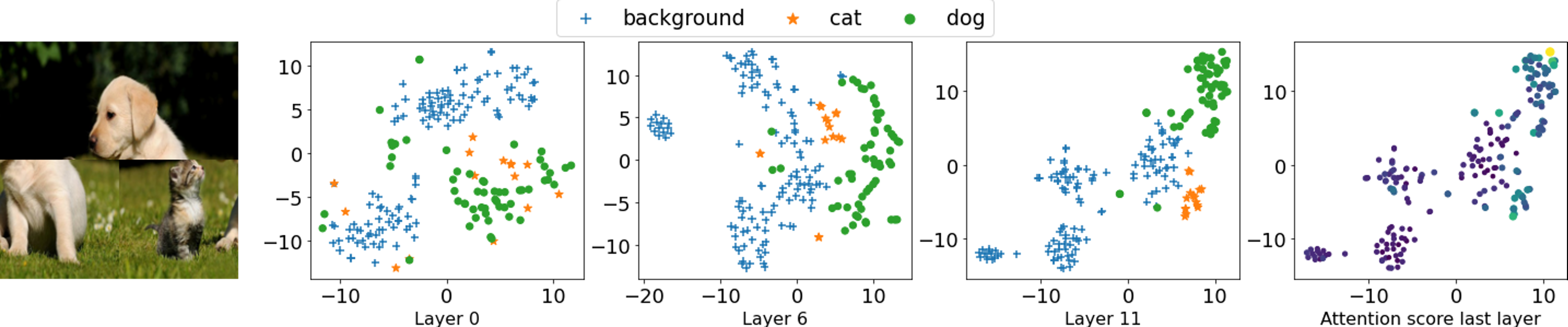}
\par\end{centering}
}\vspace{-4mm}

\subfloat[Image shuffled with the grid $8\times8$. The object embeddings do
not form clusters and attention weights for object embeddings are
still high. \label{fig:t-SNE_shuffle_8} ]{\begin{centering}
\includegraphics[width=0.95\textwidth]{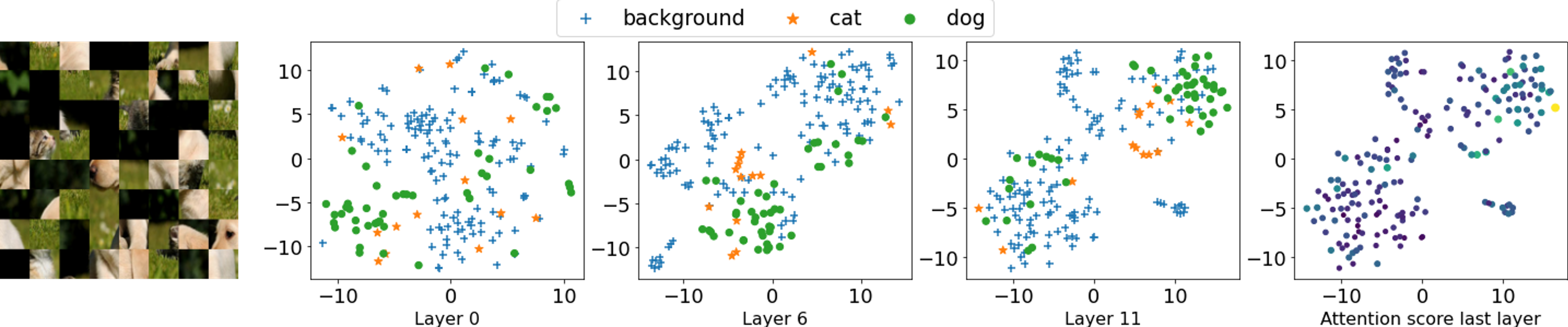}\vspace{-2mm}
\par\end{centering}
\vspace{-2mm}
}

\caption{t-SNE of feature embeddings across the layers for shuffling images.\label{fig:t-SNE_shuffle}}
\vspace{-2mm}
\end{figure}

\begin{figure}[h]
\begin{centering}
\vspace{-1mm}
\includegraphics[width=0.40\textwidth]{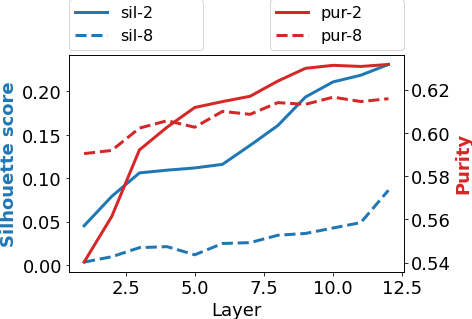} \hspace{5mm}
\includegraphics[width=0.40\textwidth]{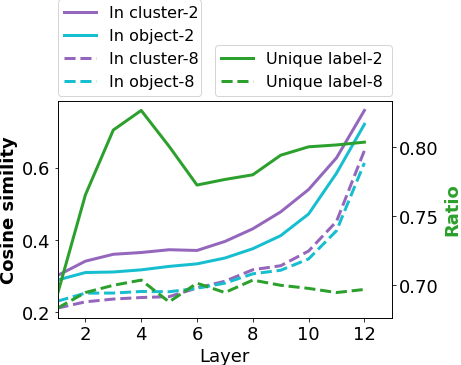}
\vspace{-2mm}
\par\end{centering}
\caption{The visualization of the average the clusteting measures for shuffling images.
The purity and Silhouette scores for the $2\times2$ shuffling are higher
than for the $8\times8$ shuffling, but they are all low. The in-cluster
and in-object similarities are rather high for the $2\times2$ shuffling,
whereas they are much lower for the $8\times8$ shuffling. \label{fig:t-SNE_embeddings-shuffe_stat}}
\vspace{-2mm}
\end{figure}
\textbf{\emph{Clustering behavior for shuffling images:}} We investigate
the clustering behavior for the patch-shuffling images. We first use
t-SNE to visualize feature embeddings across the layers for patch-shuffling
images, followed by the evaluation of the clustering measures for
the real shuffling images. We observe that for the $2\times2$ shuffling,
all measures are rather good, meaning that ViT are quite robust to
patch shuffling, whereas for the $8\times8$ shuffling, all measures are
significantly hurt.

To conclude, we notice that although in this work, we have not further exploited the clustering behavior of the feature embeddings in a real-world application, we believe this clustering behavior if strengthened properly using for example contrastive learning \cite{chen2020simple, khosla2020supervised} might possibly create more contrastive feature embeddings to benefit downstream tasks.

\section{Conclusion}

In this paper, we propose a visualization technique to visually expose the information carried in a neuron or an embedding at a certain layer of ViTs. Our visualization technique helps to visualize the local and global information on an image represented by a neuron or a group of neurons. By observing and analyzing the visual outcome of our visualization technique, we come with some findings, wherein the most important one is the clustering behavior of the embeddings across the layers of ViTs. We establish the sketch mathematical explanation for this behavior as well as conducting experiments on real images to quantitatively verify it. 
\clearpage

%%%%%%%%%%%%%%%%%%%%%%%%%%%%%%%%%%%%%%%%%%%%%%%%%%%%%%%%%%%%

\appendix

% \bibliographystyle{plain}
% \bibliography{Visualize_ViT_arxiv.bib}
% \input{Visualize_ViT_arxiv.bbl}

% \import{}{macros.tex}
\section*{Appendix}
\section{Theoretical Development}
\textbf{Mathematical explanation for the clustering behavior:} We
consider the single-head self-attention with the key matrix $W_{K}$,
the query matrix $W_{Q}$, and the value matrix $W_{V}$. We examine
the embeddings at two consecutive layers: $\mathbf{z}_{t}\in\mathbb{R}^{(N+1)\times D}$
and $\mathbf{z}_{t+1}\in\mathbb{R}^{(N+1)\times D}$, where each row
of them is an embedding. The attention weights $\mathbf{a}_{t}^{k,j}$
are computed based on the similarity of the linear projections of
$\mathbf{z}_{t}^{k}$ and $\mathbf{z}_{t}^{j}$ w.r.t. $W_{K}$ and
$W_{Q}$ respectively. We then apply the linear projection to $\mathbf{z}_{t}^{j},j=0,...,N$,
then compute $\tilde{\mathbf{z}}_{t+1}^{j},j=0,...,N$, and finally
do another linear projection to project $\tilde{\mathbf{z}}_{t+1}^{j},j=0,...,N$
back to the embedding space to obtain $\mathbf{z}_{t+1}^{j},j=0,...,N$.
To simplify our analysis, we further assume that $W_{V}=\mathbb{I}$
(i.e., the identity matrix). Therefore, we compute the embeddings
in the next layer as $\mathbf{z}_{t+1}^{j}=\sum_{k=0}^{N}\mathbf{a}_{t}^{k,j}\mathbf{z}_{t}^{k}$.
We develop the following theorem regarding the clustering behavior
of the embeddings.
\begin{thm}
Let $\text{d}\left(A\right)=\max_{\mathbf{a},\mathbf{b}\in A}\Vert\mathbf{a}-\mathbf{b}\Vert_{2}$
be the diameter of the set $A$. The following statements hold

i) $\text{d\ensuremath{\left(Z_{t+1}\right)}\ensuremath{\ensuremath{\leq\text{d}\left(Z_{t}\right)}}}$,
where $Z_{t}$ and $Z_{t+1}$ are the sets of the row vectors (i.e.,
the embeddings) of $\mathbf{z}_{t}$ and $\mathbf{z}_{t+1}$ respectively.

ii) Assume that the embeddings $Z_{t}$ (or $\mathbf{z}_{t}$) can
be grouped into $M$ clusters $Z_{t}^{m}=\left\{ \mathbf{z}_{t}^{j}:j\in G^{m}\right\} $,
where $\left[G^{m}\right]_{m=1}^{M}$ is a partition of $\left\{ 0,1,...,N\right\} $
such that if $j\in G^{m}$ and $j'\in G^{m'}$ with $m\neq m'$, the
attention weight $\epsilon_{l}<\mathbf{a}_{t}^{j',j}<\epsilon_{u}$
for $\epsilon_{u}\geq\epsilon_{l}\geq0$. This condition means that
if $\mathbf{z}_{t}^{j}$ and $\mathbf{z}_{t}^{j'}$ belong to two
different clusters, they are less similar and their attention weight
$\mathbf{a}_{t}^{j',j}$ is possibly small. Let $N_{m}=\left|G_{m}\right|$
be the cardinality of the cluster $m$. For any $m$, we have the
following inequality:
\begin{align}
\text{d}\left(Z_{t+1}^{m}\right) & <\left(1-A_{m}\epsilon_{l}\right)^{2}\text{d}\left(Z_{t}^{m}\right)+A_{m}\epsilon_{u}\text{d}\left(Z_{t}\right)\left(A_{m}\epsilon_{u}+2\right),
\end{align}
where $A_m= N+1-N_m$ and $Z_{t+1}^{m}=\left\{ \mathbf{z}_{t+1}^{i}:i\in G^{m}\right\} $
is the cluster in the next layer. Note that $\text{d}\left(Z_{t+1}^{m}\right)$
and $\text{d}\left(Z_{t}^{m}\right)$ are exactly the intra-cluster
distances for these clusters.
\end{thm}

\textbf{Proof of Theorem \ref{thm:intra_distance}.}

i) Because the distance $h(\mathbf{a},\mathbf{b})=\Vert\mathbf{a}-\mathbf{b}\Vert_{2}$
is a convex function, we have
\begin{align*}
\norm{\mathbf{z}_{t}^{j}-\mathbf{z}_{t}^{j'}}_{2} & =\norm{\sum_{k=0}^{N}\mathbf{a}_{t}^{k,j}\mathbf{z}_{t}^{k}-\sum_{k=0}^{N}\mathbf{a}_{t}^{k,j'}\mathbf{z}_{t}^{k}}_{2}=h\left(\sum_{k=0}^{N}\mathbf{a}_{t}^{k,j}\mathbf{z}_{t}^{k},\sum_{k=0}^{N}\mathbf{a}_{t}^{k,j'}\mathbf{z}_{t}^{k}\right)\\
\leq & \sum_{k=0}^{N}\sum_{k'=0}^{N}\mathbf{a}_{t}^{k,j}\mathbf{a}_{t}^{k',j'}h\left(\mathbf{z}_{t}^{k},\mathbf{z}_{t}^{k'}\right)\leq\sum_{k=0}^{N}\sum_{k'=0}^{N}\mathbf{a}_{t}^{k,j}\mathbf{a}_{t}^{k',j'}\text{d}\left(Z_{t}\right)\\
= & \sum_{k=0}^{N}\mathbf{a}_{t}^{k,j}\sum_{k'=0}^{N}\mathbf{a}_{t}^{k',j'}\text{d}\left(Z_{t}\right)=\text{d}\left(Z_{t}\right).
\end{align*}

Therefore, we reach $\text{d\ensuremath{\left(Z_{t+1}\right)}\ensuremath{\ensuremath{\leq\text{d}\left(Z_{t}\right)}}}$.

ii) For any $j,j'\in G^{m}$, we derive as 
\begin{align*}
\norm{\mathbf{z}_{t}^{j}-\mathbf{z}_{t}^{j'}}_{2}& =\norm{\sum_{k=0}^{N}\mathbf{a}_{t}^{k,j}\mathbf{z}_{t}^{k}-\sum_{k=0}^{N}\mathbf{a}_{t}^{k,j'}\mathbf{z}_{t}^{k}}_{2}=h\left(\sum_{k=0}^{N}\mathbf{a}_{t}^{k,j}\mathbf{z}_{t}^{k},\sum_{k=0}^{N}\mathbf{a}_{t}^{k,j'}\mathbf{z}_{t}^{k}\right)\\
 & \leq\sum_{k=0}^{N}\sum_{k'=0}^{N}\mathbf{a}_{t}^{k,j}\mathbf{a}_{t}^{k',j'}h\left(\mathbf{z}_{t}^{k},\mathbf{z}_{t}^{k'}\right)\\
 & \leq\sum_{k\in G^{m}}\sum_{k'\in G^{m}}\mathbf{a}_{t}^{k,j}\mathbf{a}_{t}^{k',j'}h\left(\mathbf{z}_{t}^{k},\mathbf{z}_{t}^{k'}\right)+\sum_{k\in G^{m}}\sum_{k'\notin G^{m}}\mathbf{a}_{t}^{k,j}\mathbf{a}_{t}^{k',j'}h\left(\mathbf{z}_{t}^{k},\mathbf{z}_{t}^{k'}\right)\\
 & +\sum_{k\notin G^{m}}\sum_{k'\in G^{m}}\mathbf{a}_{t}^{k,j}\mathbf{a}_{t}^{k',j'}h\left(\mathbf{z}_{t}^{k},\mathbf{z}_{t}^{k'}\right)+\sum_{k\notin G^{m}}\sum_{k'\notin G^{m}}\mathbf{a}_{t}^{k,j}\mathbf{a}_{t}^{k',j'}h\left(\mathbf{z}_{t}^{k},\mathbf{z}_{t}^{k'}\right)\\
 & \leq\sum_{k\in G^{m}}\sum_{k'\in G^{m}}\mathbf{a}_{t}^{k,j}\mathbf{a}_{t}^{k',j'}\text{d}\left(Z_{t}^{m}\right)+\sum_{k\in G^{m}}\sum_{k'\notin G^{m}}\mathbf{a}_{t}^{k,j}\mathbf{a}_{t}^{k',j'}\text{d}\left(Z_{t}\right)\\
 & +\sum_{k\notin G^{m}}\sum_{k'\in G^{m}}\mathbf{a}_{t}^{k,j}\mathbf{a}_{t}^{k',j'}\text{d}\left(Z_{t}\right)+\sum_{k\notin G^{m}}\sum_{k'\notin G^{m}}\mathbf{a}_{t}^{k,j}\mathbf{a}_{t}^{k',j'}\text{d}\left(Z_{t}\right).
\end{align*}

We further manipulate as
\begin{align*}
\sum_{k\in G^{m}}\sum_{k'\in G^{m}}\mathbf{a}_{t}^{k,j}\mathbf{a}_{t}^{k',j'}\text{d}\left(Z_{t}^{m}\right) & =\text{d}\left(Z_{t}^{m}\right)\sum_{k\in G^{m}}\mathbf{a}_{t}^{k,j}\sum_{k'\in G^{m}}\mathbf{a}_{t}^{k',j'}\\
= & \text{d}\left(Z_{t}^{m}\right)\left(1-\sum_{k\notin G^{m}}\mathbf{a}_{t}^{k,j}\right)\left(1-\sum_{k'\notin G^{m}}\mathbf{a}_{t}^{k',j'}\right)\\
< & \text{d}\left(Z_{t}^{m}\right)\left(1-A_{m}\epsilon_{l}\right)^{2}\text{d}\left(Z_{t}^{m}\right).
\end{align*}
\begin{align*}
\sum_{k\in G^{m}}\sum_{k'\notin G^{m}}\mathbf{a}_{t}^{k,j}\mathbf{a}_{t}^{k',j'}\text{d}\left(Z_{t}\right) & =\text{d}\left(Z_{t}\right)\sum_{k\in G^{m}}\mathbf{a}_{t}^{k,j}\sum_{k'\notin G^{m}}\mathbf{a}_{t}^{k',j'}\\
 & <A_{m}\epsilon_{u}\text{d}\left(Z_{t}\right).
\end{align*}

\begin{align*}
\sum_{k\notin G^{m}}\sum_{k'\in G^{m}}\mathbf{a}_{t}^{k,j}\mathbf{a}_{t}^{k',j'}\text{d}\left(Z_{t}\right) & =\text{d}\left(Z_{t}\right)\sum_{k\notin G^{m}}\mathbf{a}_{t}^{k,j}\sum_{k'\in G^{m}}\mathbf{a}_{t}^{k',j'}\\
< & A_{m}\epsilon_{u}\text{d}\left(Z_{t}\right).
\end{align*}
\begin{align*}
\sum_{k\notin G^{m}}\sum_{k'\notin G^{m}}\mathbf{a}_{t}^{k,j}\mathbf{a}_{t}^{k',j'}\text{d}\left(Z_{t}\right) & =\text{d}\left(Z_{t}\right)\sum_{k\notin G^{m}}\mathbf{a}_{t}^{k,j}\sum_{k'\notin G^{m}}\mathbf{a}_{t}^{k',j'}\\
< & A_{m}^{2}\epsilon_{u}^{2}\text{d}\left(Z_{t}\right).
\end{align*}

Finally, we arrive at
\[
\norm{\mathbf{z}_{t}^{j}-\mathbf{z}_{t}^{j'}}_{2}^{2}<\left(1-A_{m}\epsilon_{l}\right)^{2}\text{d}\left(Z_{t}^{m}\right)+A_{m}\epsilon_{u}\text{d}\left(Z_{t}\right)\left(A_{m}\epsilon_{u}+2\right).
\]

\begin{align*}
\text{d}\left(Z_{t+1}^{m}\right) & <\left(1-A_{m}\epsilon_{l}\right)^{2}\text{d}\left(Z_{t}^{m}\right)+A_{m}\epsilon_{u}\text{d}\left(Z_{t}\right)\left(A_{m}\epsilon_{u}+2\right).
\end{align*}

It is worth noting that our proof can be generalized to any convex
distances, e.g., $h(\mathbf{a},\mathbf{b})=\Vert\mathbf{a}-\mathbf{b}\Vert_{p}$ ($p \geq 1$). We made a minor typo in Inequality (\ref{eq:distance_reduce}) in the main paper. We will correct it in the revised version. Sorry for this inconvenience.

\section{Neural behavior for image occlusion}
\subsection{Clustering Metrics}

\paragraph{Purity score \cite{pml1Book}: }

Given a cluster $Z_{t}^{m}$, we define $p_{m}=\max_{i}p_{mi}$ as
the frequency of the most dominated labels (e.g., object types or
background) and compute the purity as:
\[
\text{purity}=\sum_{m=1}^{M}p_{m}\frac{N_{m}}{N+1}.
\]

A \emph{high purity} means that each cluster is more pure or mostly
dominated by a label.

\paragraph{Silhouette score \cite{silhouette}:}

For each $\mathbf{z}_{t}^{j}\in Z_{t}^{m}$, we compute the corresponding
intra-distance and inter-distance as
\[
a(\mathbf{z}_{t}^{j})=\frac{1}{\left|N_{m}\right|-1}\sum_{j'\in G^{m},j'\neq j}\norm{\mathbf{z}_{t}^{j}-\mathbf{z}_{t}^{j'}}_{2}.
\]
\[
b(\mathbf{z}_{t}^{j})=\min_{m'\neq m}\frac{1}{\left|N_{m}^{'}\right|}\sum_{j'\in G^{m'}}\norm{\norm{\mathbf{z}_{t}^{j}-\mathbf{z}_{t}^{j'}}_{2}}.
\]

The Silhouette score at $\mathbf{z}_{t}^{j}\in Z_{t}^{m}$ is computed
as 
\[
\text{Silhouette}\left(\mathbf{z}_{t}^{j}\right)=\frac{b(\mathbf{z}_{t}^{j})-a(\mathbf{z}_{t}^{j})}{\max\left\{ a(\mathbf{z}_{t}^{j}),b(\mathbf{z}_{t}^{j})\right\} },
\]
and the Silhouette score is the average of the Silhouette scores at
the embeddings. 

A \emph{high Silhouette} score implies the intra-cluster distances
are relatively smaller than the inter-cluster distances.

\paragraph{Unique label ratio:}

This is measured as the ratio of the number of unique dominated object-type
labels and the number of unique object-type labels. A \emph{high unique
label ratio} implies that for most of object-type label, there exists
at least a cluster dominated by this object-type label.

We investigate the clustering behavior for image occlusion. In which, we apply three types of occlusion: \textit{Random drop}, \textit{salient} and \textit{non-salient} occlusion, with different ratios (from $0.2$ to $0.8$ for random drop and salient occlusion, from $0.2$ to $1.0$ for non-salient occlusion). Note that the number of data for each setting is different because we select only images with at least two big objects as described in Section 3.2. Additionally, patches in the images are dropped randomly, therefore, possibly reducing the number of tokens for each object in the image. 
This process is mentioned in detail below.

\subsection{Random drop} \label{subsec:random_drop}

\begin{figure}
\vspace{-2mm}

\subfloat[Random drop at ratio 0.2.\label{fig:t-SNE_ran0.2}]{\begin{centering}
\includegraphics[width=0.98\textwidth]{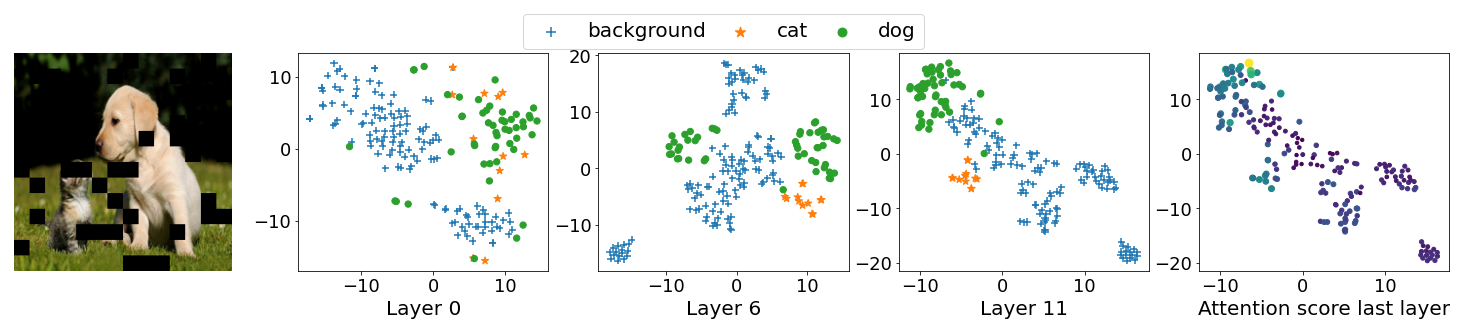}
\par\end{centering}
}\vspace{-2mm}

\subfloat[Random drop at ratio 0.4.\label{fig:t-SNE_ran0.4}]{\begin{centering}
\includegraphics[width=0.98\textwidth]{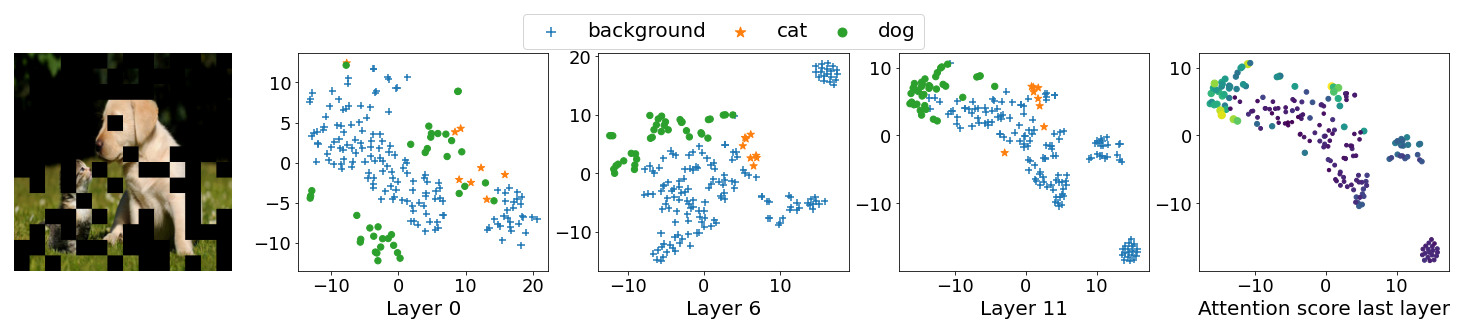}
\par\end{centering}
}\vspace{-2mm}

\subfloat[Random drop at ratio 0.6.\label{fig:t-SNE_ran0.6}]{\begin{centering}
\includegraphics[width=0.98\textwidth]{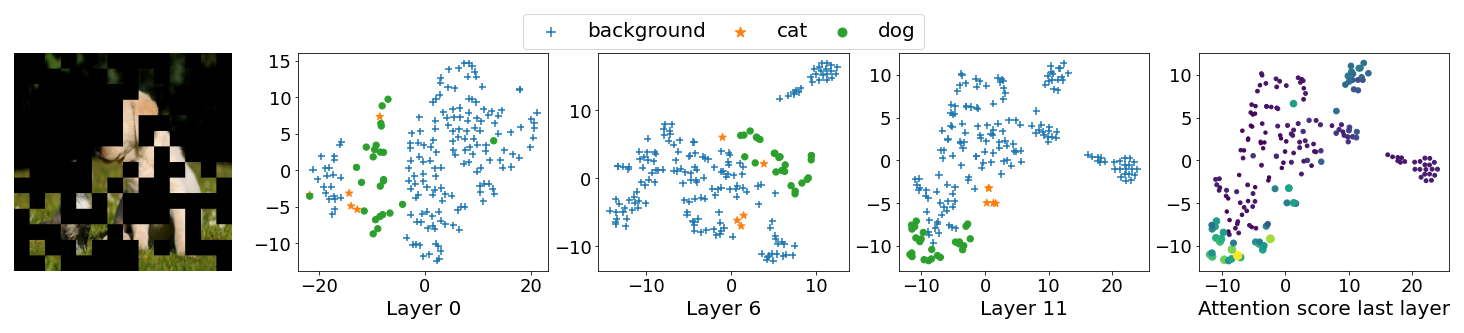}
\par\end{centering}
}\vspace{-2mm}

\subfloat[Random drop at ratio 0.8.\label{fig:t-SNE_ran0.8}]{\begin{centering}
\includegraphics[width=0.98\textwidth]{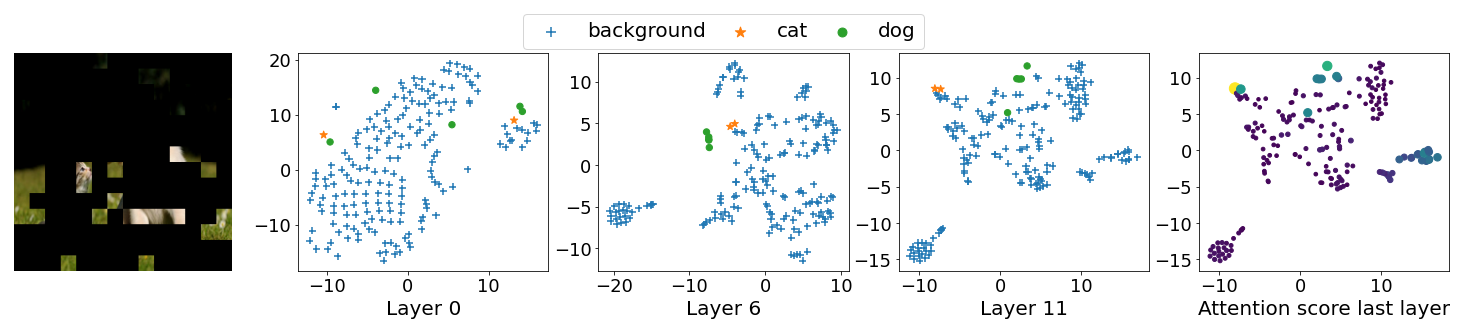}
\par\end{centering}
}\vspace{-2mm}

\caption{t-SNE of feature embeddings across the layers of ViT. The Pascal VOC dataset is processed with random drop at different ratio from $0.2$ to $0.8$ \label{fig:t-SNE_ran}}

\centering{}\vspace{-6mm}
\end{figure}

\begin{figure}
\includegraphics[width=0.98\textwidth]{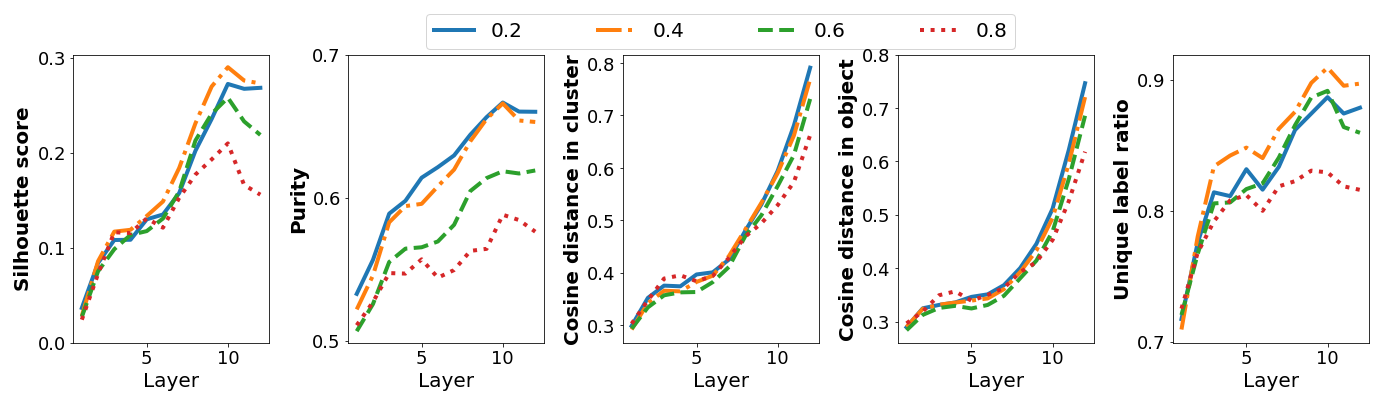}
\vspace{-2mm}
\caption{Clustering measurement of embeddings with \textbf{random drop}. The images in Pascal VOC dataset are processed with random dropping at different ratio from $0.2$ to $0.8$. With higher ratio of occlusion, the ViTs is getting harder to distinguish object to object and also object to non-object \label{fig:stat_ran}}

\centering{}
\end{figure}

We randomly choose $k$ patches in each image to change pixels to $0$, called un-informative patches, with $k = n*r$, $n$ is the total number of patch in the image and $r \in \{0.2, 0.4, 0.6, 0.8\}$. When $r$ increases, the number of informative patches decreases as same as the number of object tokens. As mentioned in Section 3.2 in the main paper, we select the images with at least two big objects to investigate the clustering behavior, so that the number of data matching the requirement decreases when $r$ increases. Particularly, there are $494$, $438$, $337$, $177$ valid images for corresponding $r$. A sample for each value of $r$ is displayed in Figure \ref{fig:t-SNE_ran} along with its t-SNE of feature embeddings. It can be observed that with $r = \{0.2, 0.4, 0.6\}$, ViT is still able to group the object tokens into clusters at higher layers. But increasing r to 0.8, ViT barely distinguishes the object and non-object tokens. However, in all settings of $r$, the feature embeddings belonging to the same object still stay closer together and the attention weights for object-patch embeddings are also higher than the others. 

Figure \ref{fig:stat_ran} compares the clustering measures of random drop at different ratios. In general, all scores decrease when $r$ increases but they all increase across layers. This demonstrates the ability of ViTs to group embeddings of the same objects and find the correct object patches to pay attention to. The purity and unique label ratio drop across values of $r$ shows ViTs could be noise by background when losing significant number of informative patches. Mean that ViTs find it harder to distinguish background and foreground if the information is not enough.

\begin{figure}
\vspace{-2mm}

\subfloat[Salient occlusion at ratio 0.2.\label{fig:t-SNE_sal0.2}]{\begin{centering}
\includegraphics[width=0.98\textwidth]{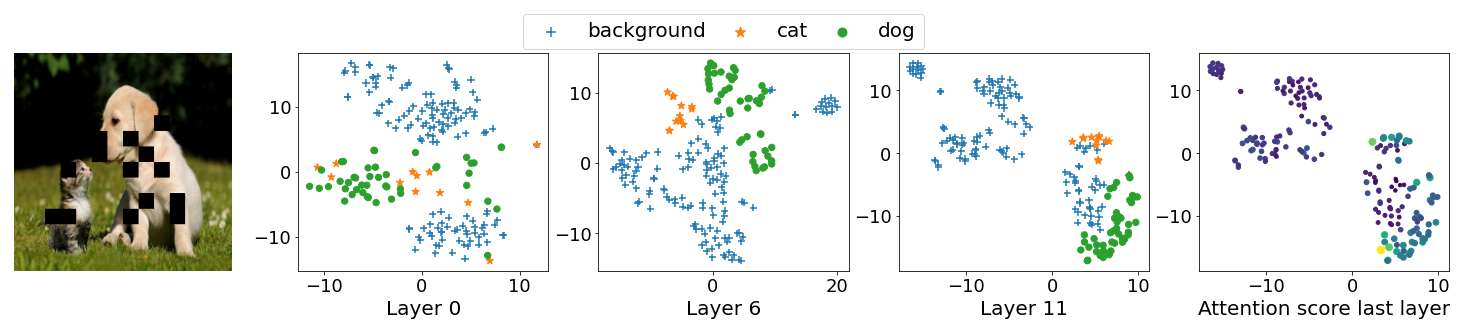}
\par\end{centering}
}\vspace{-2mm}

\subfloat[Salient occlusion at ratio 0.4.\label{fig:t-SNE_sal0.4}]{\begin{centering}
\includegraphics[width=0.98\textwidth]{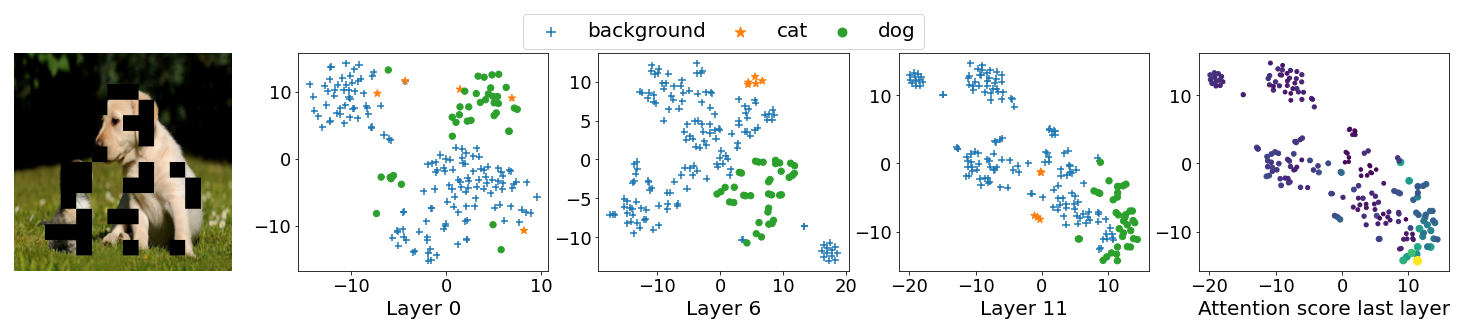}
\par\end{centering}
}\vspace{-2mm}
\ContinuedFloat

\subfloat[Salient occlusion at ratio 0.6.\label{fig:t-SNE_sal0.6}]{\begin{centering}
\includegraphics[width=0.98\textwidth]{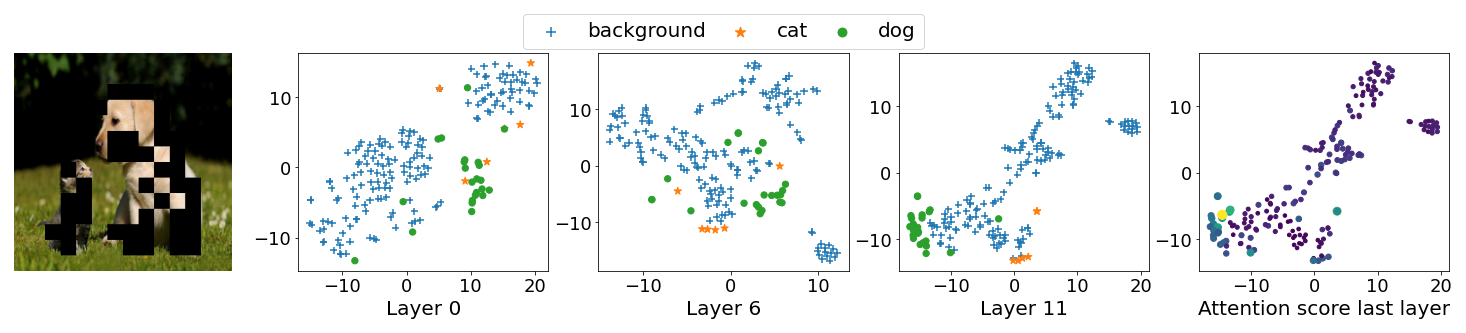}
\par\end{centering}
}\vspace{-2mm}

\subfloat[Salient occlusion at ratio 0.8.\label{fig:t-SNE_sal0.8}]{\begin{centering}
\includegraphics[width=0.98\textwidth]{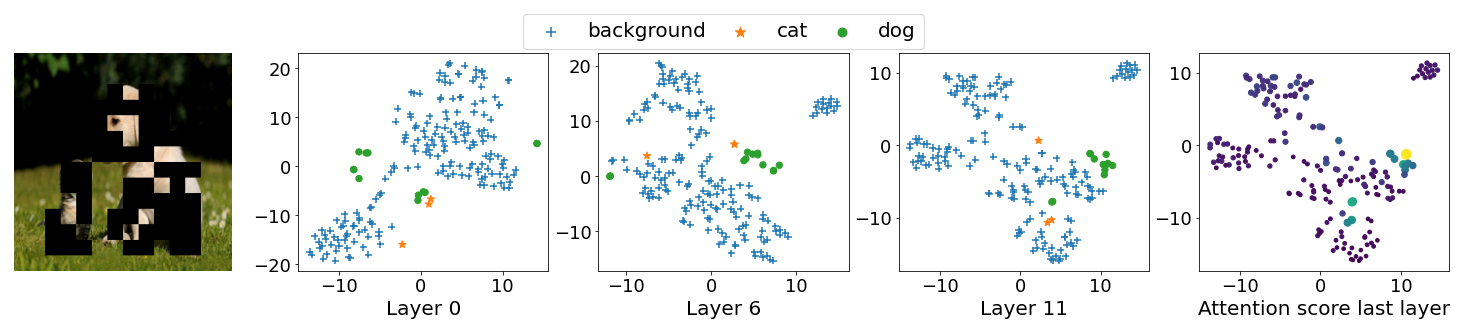}
\par\end{centering}
}\vspace{-2mm}

\caption{t-SNE of feature embeddings across the layers of ViT. The Pascal VOC dataset is processed with salient occlusion at different ratio from $0.2$ to $0.8$ \label{fig:t-SNE_sal}}

\centering{}\vspace{-6mm}
\end{figure}

\begin{figure}
\includegraphics[width=0.98\textwidth]{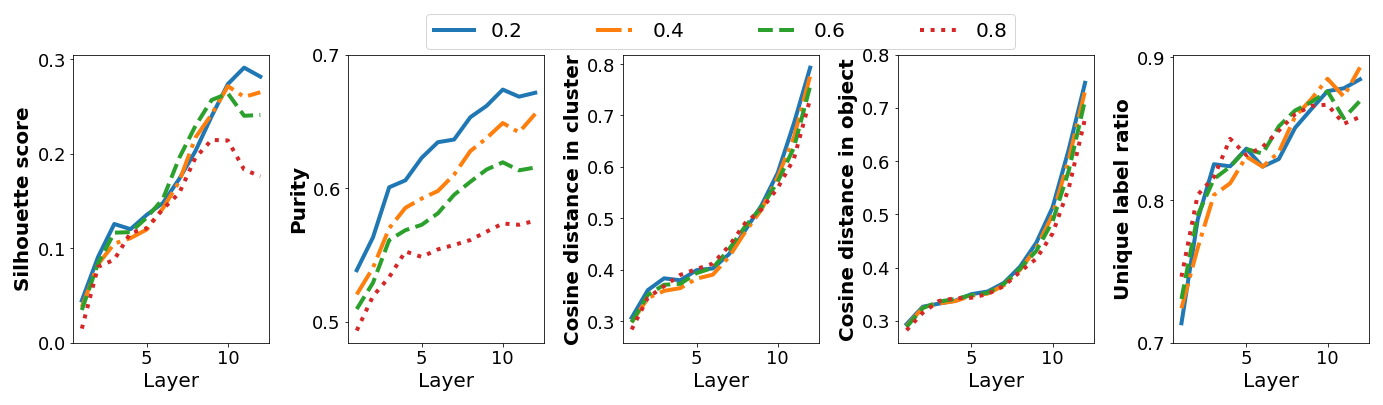}

\caption{Clustering measurement of embeddings with \textbf{salient occlusion}. The images in Pascal VOC dataset are processed with salient occlusion at different ratio from $0.2$ to $0.8$. With higher ratio of occlusion, embeddings in cluster or of an object still preserve the high similarity but ViTs find it harder to distinguish object to object. \label{fig:stat_sal}}

\centering{}
\end{figure}
% \clearpage

\subsection{Salient occlusion} \label{subsec:sal_occ}

Instead of randomly choosing a part of all patches in an image, the salient occlusion only drops some object patches, depending on ratio $r$. So that, even with the same value of $r$, the number of valid images is different between random drop and salient occlusion. Particularly, $494$, $441$, $355$, $194$ valid images for $r \in \{0.2, 0.4, 0.6, 0.8\}$ respectively. A sample for each value of $r$ is presented in Figure \ref{fig:t-SNE_sal}. The behavior of feature embeddings is also similar to random drop occlusion that increasing $r$, ViTs find it harder to distinguish object and non-object embeddings but the cosine similarity of object embeddings are higher across layers.

Figure \ref{fig:stat_sal} shows the clustering score for ViT when apply the salient occlusion process. The purity score is the most significant change along the increase of $r$. Means that it gets harder to separate embeddings of different objects when gradually losing object detail (object patches). Together with the modest change of the unique label ratio across values of $r$ show ViTs may learn to separate objects in different categories (i.e. object cat and object dog) better than in the same categories (i.e. two object cats).

\subsection{Non-salient occlusion} \label{subsec:non_sal_occ}

\begin{figure}
% \ContinuedFloat
\vspace{-2mm}

\subfloat[Non-salient occlusion at ratio 0.2.\label{fig:t-SNE_nosal0.2}]{\begin{centering}
\includegraphics[width=0.98\textwidth]{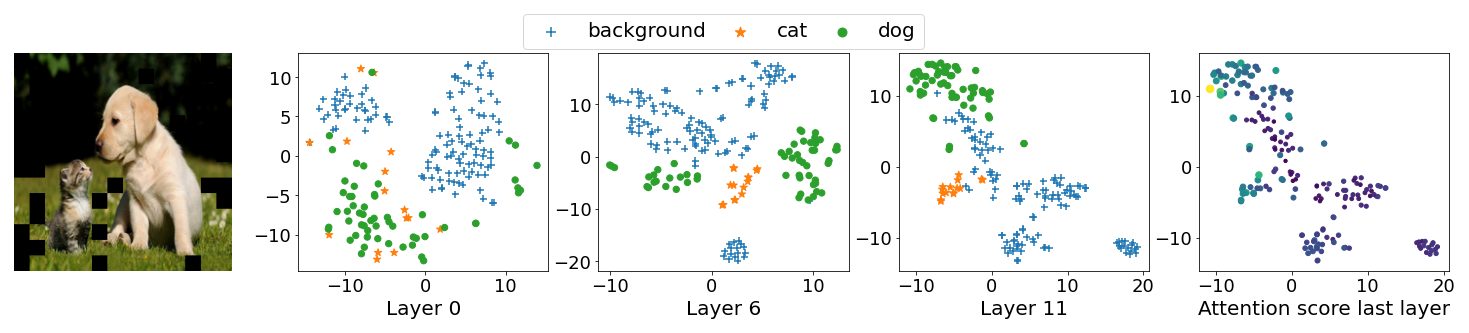}
\par\end{centering}
}\vspace{-2mm}

\subfloat[Non-salient occlusion at ratio 0.4.\label{fig:t-SNE_nosal0.4}]{\begin{centering}
\includegraphics[width=0.98\textwidth]{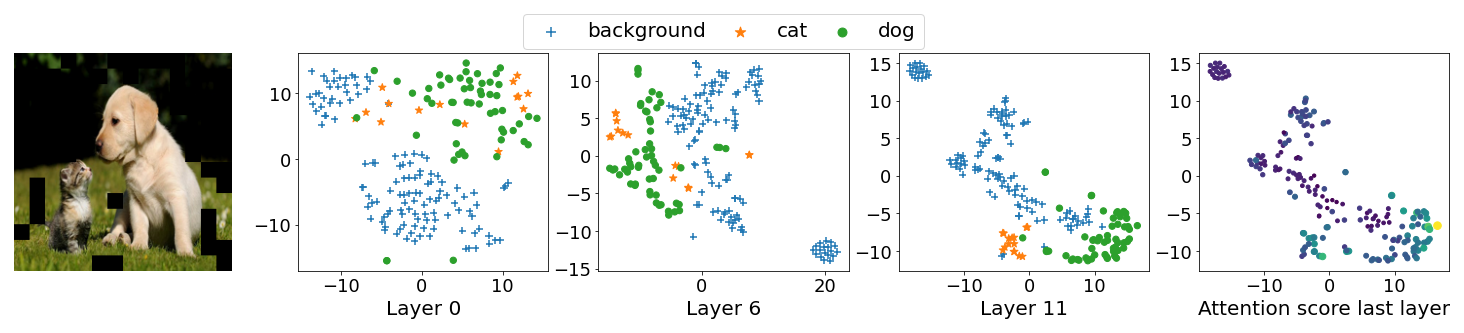}
\par\end{centering}
}\vspace{-2mm}

\subfloat[Non-salient occlusion at ratio 0.6.\label{fig:t-SNE_nosal0.6}]{\begin{centering}
\includegraphics[width=0.98\textwidth]{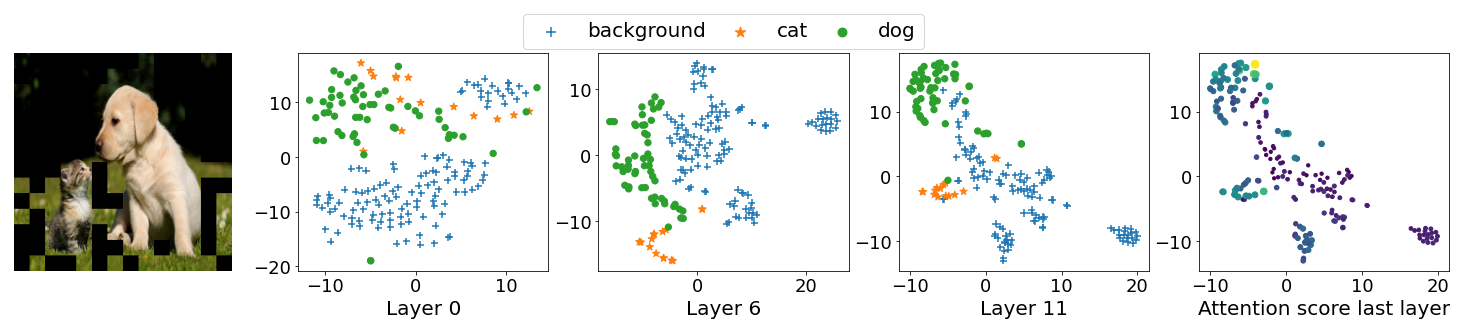}
\par\end{centering}
}\vspace{-2mm}

\subfloat[Non-salient occlusion at ratio 0.8.\label{fig:t-SNE_nosal0.8}]{\begin{centering}
\includegraphics[width=0.98\textwidth]{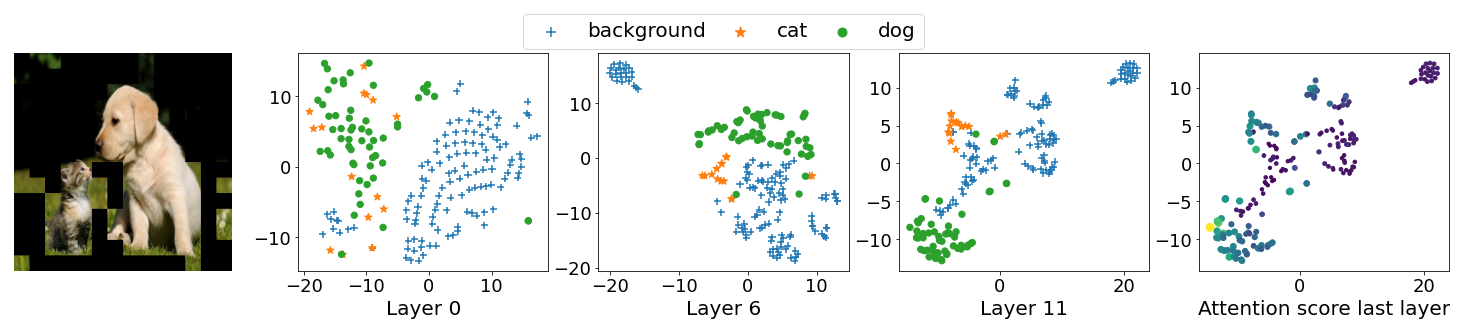}
\par\end{centering}
}\vspace{-2mm}

\subfloat[Non-salient occlusion at ratio 1.0.\label{fig:t-SNE_nosal1}]{\begin{centering}
\includegraphics[width=0.98\textwidth]{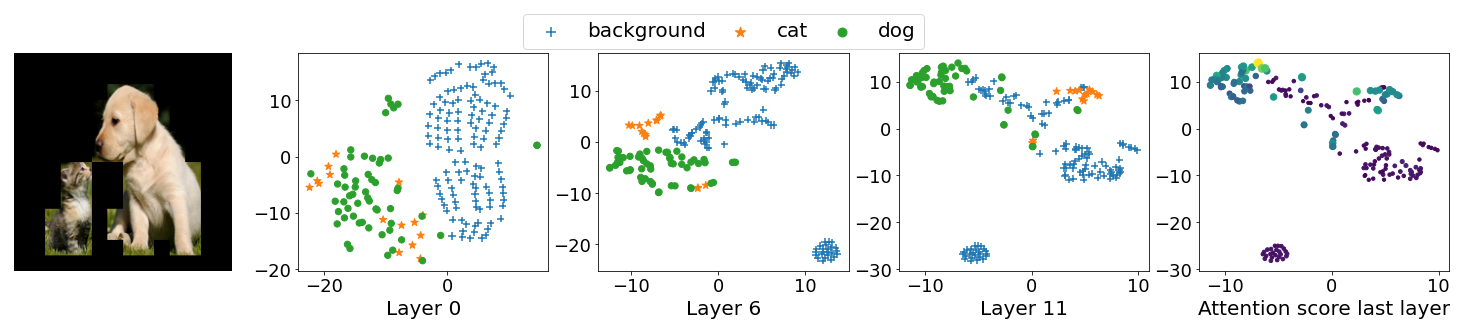}
\par\end{centering}
}\vspace{-2mm}

\caption{t-SNE of feature embeddings across the layers of ViT. The Pascal VOC dataset is processed with \textbf{non-salient occlusion} at different ratio from $0.2$ to $0.8$ \label{fig:t-SNE_nosal}}

\centering{}\vspace{-6mm}
\end{figure}

\begin{figure}[!ht]
\includegraphics[width=0.98\textwidth]{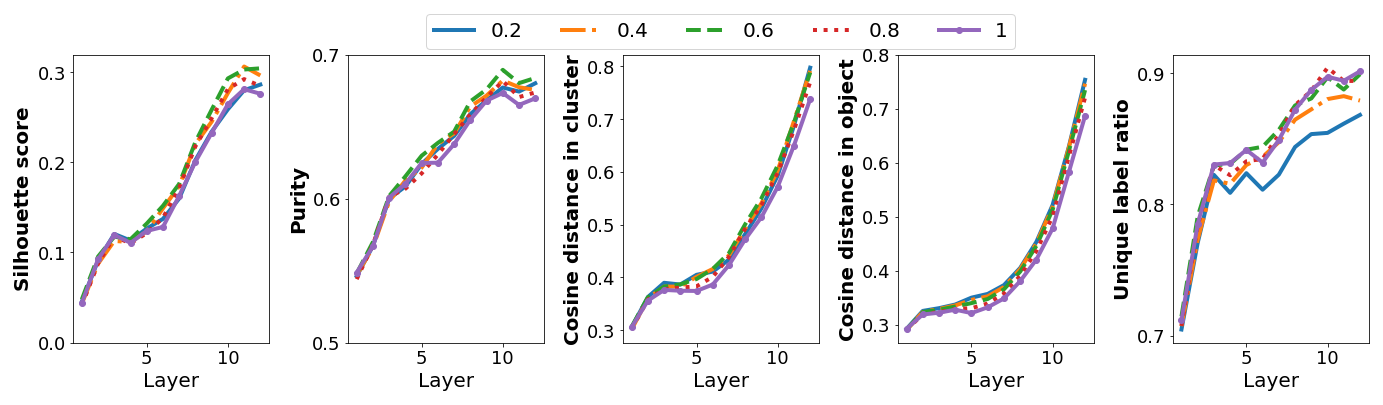}

\caption{Clustering measurement of embeddings with \textbf{non-salient occlusion}. The images in Pascal VOC dataset are processed with non-salient occlusion at different ratio from $0.2$ to $0.8$. \label{fig:stat_nosal}}

\centering{}
\end{figure}

The pre-process is the same as salient occlusion but choosing the non-object patches to drop. As mentioned above, we select the images with big objects so that the non-salient occlusion does not impact to number of valid data, which is the same $530$ images for all value of $r$. Samples for different ratios are presented in Figure \ref{fig:t-SNE_nosal} while the clustering measures in presented in Figure \ref{fig:stat_nosal}. Apparently, increasing $r$ helps remove background noise, therefore, ViTs could group object embeddings clearly in the very first layer. It also shows this phenomenon in unique label ratio plot since more objects are formed in clusters increasing ratio $r$. The other cluster measures in Figure \ref{fig:stat_nosal} shows a small change. In general, removing background makes ViTs more focus on distinguish objects and object embeddings less attractive by non-object ones.

\begin{figure}[!ht]
\includegraphics[width=0.98\textwidth]{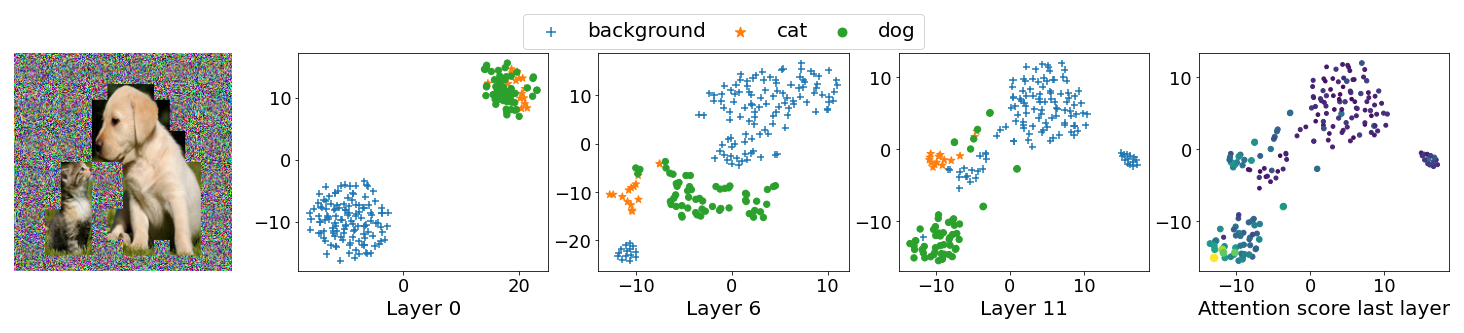}

\caption{A sample of feature embeddings when replacing all pixels in non-object patches with random noise (instead of changing to $0$ like non-salient occlusion) \label{fig:t-SNE_nosal_nonzero}}

\centering{}
\end{figure}

Interestingly, as presented in Figure \ref{fig:t-SNE_nosal1}, non-object embeddings are formed in one cluster in very first layers (as the value of these patches is all zeros). However, in higher layers (e.g., 6th and 11th layers), we clearly see two clusters of embeddings coressponding to non-salient patches. We reason that the position of patches affects to embedding across ViT's layers. When non-salient patches are set to zero, the positional encodings are the only factor impact the attention scores, so that the attention scores to object-patches might be higher for non-salient patches, which are close to object-patches. As a result, the embeddings corresponding to non-salient patches, which are close to object-patches, might gather more information in comparison with the ones that are far way from object-patches, leading to two separate clusters in higher layers (as discussion in Section 3.1.2 in the main paper, the visualization of higher layer linearly aggregates information from lower layer weighted by attention score). To confirm this observation, we change value of non-salient patches to random noise instead of zero, which would reduce the impact of positional encoding when calculating attention map. It can be seen from Figure \ref{fig:t-SNE_nosal_nonzero} that tokens corresponding to non-salient patches are well grouped in only one cluster.

% Additionally, the position of patches could also show its effect to embedding across ViT's layers that can be observed in Figure \ref{fig:t-SNE_nosal1} when we change all pixels in none-object patches to $0$. However, when we change them to random colors, the position embeddings show less impact and ViTs could separate object and non-object patch embeddings even better (Figure \ref{fig:t-SNE_nosal_nonzero}).

\end{document}